\documentclass[final,3p,times,authoryear]{elsarticle}
\usepackage{natbib}
\usepackage{amssymb}
\usepackage{amsmath}
\usepackage{amsthm}
\usepackage{algorithm}
\usepackage{algpseudocode}
\usepackage{booktabs}
\usepackage{siunitx}
\usepackage{subcaption}
\usepackage{caption}
\usepackage{graphicx}
\usepackage{xr}
\usepackage{float}
\usepackage{hyperref}
\usepackage{cleveref}

\begin{document}

\begin{frontmatter}



\title{An Explainable AI-based approach for Monitoring Animal Health}


\author[awadh,csio]{Rahul Jana}

\author[awadh,csio]{Shubham Dixit} 

\author[csio,AcSIR,DHE]{Mrityunjay Sharma} 

\author[csio,AcSIR]{Ritesh Kumar} 

\affiliation[awadh]{organization={TIF - Agriculture and Water Technology Development Hub(AWaDH), Indian Institute of Technology},
            city={Bara Phool},
            postcode={140001}, 
            state={Punjab},
            country={India}}
            
\affiliation[csio]{organization={CSIR-Central Scientific Instruments Organization},
            city={Chandigarh},
            postcode={160030}, 
            country={India}}

\affiliation[AcSIR]{organization={Academy of Scientific and Innovative Research (AcSIR)},
            city={Ghaziabad},
            postcode={201002}, 
            state={Uttar Pradesh},
            country={India}}

\affiliation[DHE]{organization={Department of Higher Education},
            city={Shimla},
            postcode={171001}, 
            state={Himachal Pradesh},
            country={India}}

\begin{abstract}
Monitoring cattle health and optimizing yield are key challenges faced by dairy farmers due to difficulties in tracking all animals on the farm. This work aims to showcase modern data-driven farming practices based on explainable machine learning(ML) methods that explain the activity and behavior of dairy cattle (cows). Continuous data collection of 3-axis accelerometer sensors and usage of robust ML methodologies and algorithms, provide farmers and researchers with actionable information on cattle's activity, allowing farmers to make informed decisions and incorporate sustainable practices. This study utilizes Bluetooth-based Internet-of-Things (IoT) devices and 4G networks for seamless data transmission, immediate analysis, inference generation, and explains the models' performance with explainability frameworks. Special emphasis is put on the pre-processing of the accelerometers' time series data, including the extraction of statistical characteristics, signal processing techniques, and lag-based features using the sliding-window technique. Various hyperparameter-optimized ML models are evaluated across varying window lengths for activity classification. The k-nearest neighbor Classifier achieved the best performance, with AUC of 0.98 $\pm$ 0.0026 on the training set and 0.99 on testing set). In order to ensure transparency, Explainable AI based frameworks such as SHAP is used to interpret feature importance that can be understood and used by practitioners. A detailed comparison of the important features, along with the stability analysis of selected features, supports development of explainable and practical ML models for sustainable livestock management.\end{abstract}




\end{frontmatter}



\section{Introduction}
India is ranked first in milk production, contributing 24.76\% of global milk production. Milk production has grown at a Compound Annual Growth Rate (CAGR) of 5.62\% over the past 10 years, from 146.31 million tonnes(2014-15) to 239.30 million tonnes in the year 2023-24 \citep{pib2025}. In the year 2021-22, this contributed 5\% of the national economy. However, farmers face massive challenges in terms of the efficacy, quality and quantity of milk production. Milk production in dairy animals depends on overall health, frequency of insemination, and the calving. To better understand and manage these factors, 3-axis accelerometers have been utilized to classify animals' behaviors, such as resting, ruminating, walking \citep{williams2017application}, and monitoring their movements and activities \citep{neethirajan2020role}. Accelerometers placed on collars, legs, and ear tags can classify and analyze animal \citep{rahman2018cattle,gonzalez2015behavioral,ruuska2016validation} behavior. Statistical features have become important for accurately categorizing and predicting cow behaviors. Frequency domain statistical features such as mean, standard deviation, skewness, kurtosis, root mean square, median, etc, are used to classify \citep{arablouei2021situ,tran2021iot} the cattle behaviors. These approaches showcase the specific events and capture nuances in cow behavior. For livestock management, the RandomForest Model \citep{vazquez2015classification} has also been utilized to classify dairy cow behaviors based on leg-mounted triaxial accelerometer data. The sensor's data is then split into a 16-second window of data with a sample rate of 1 sample/second to extract the features such as mean, standard deviation, root mean square, median, and range from the acceleration data to achieve high classification performance. Several studies have investigated the impact of window length on the accuracy of activity recognition \citep{decandia2018effect,fida2015varying,neirurerova2021effect}. The window length method is used for segmenting accelerometer data into smaller time intervals, allowing for the extraction of features from each segment. The optimal window length varies depending on the behavior being classified, and its selection is crucial for accurate activity recognition. Locomotion features such as mean stride length and frequency offer valuable insights into the health status of dairy cows. \citep{neirurerova2021effect} and activity levels. Based on this, accelerometer data and ML have been used in various studies to classify animal behaviors of interest. A deep learning model has been used for acquiring data by RGB-D camera to measure individual cow feed intake \citep{bezen2020computer}. Decision tree algorithm, logistic regression, and multi-class support vector machines (SVMs) have been used on the tri-axial acceleration data of housed dairy cows \citep{vazquez2015classification,martiskainen2009cow,robert2009evaluation}. Optimally tuned extreme Gradient Boosting (XGBoost) and Random Forests classifiers\citep{dutta2022moonitor} perform well to classify activities such as standing, lying, standing and ruminating, lying and ruminating, walking, and grazing, etc.
The integration of multiple sensors and machine learning algorithms have also proven effective in detecting cow diseases. For instance, feeding, lying, standing, and walking behaviors can be successfully classified using back-mounted pressure sensors in conjunction with the SVM model\citep{unold2020iot,riaboff2022predicting}.
\noindent The existing methods applied by previous researchers have largely overlooked explainability,, we  focus specifically on implementing explainability and understanding the interaction of various features derived from the raw dataset. This study will help in understanding and interpreting how feature values change in relation to cow movement. Exhaustive set of features, such as statistical features, wavelet features, etc, are derived using a custom-built feature extraction pipeline.  We also explored multiple window sizes during feature extraction and evaluated model performance across these variations.

\section{Methodology}
\label{Methodology}


\subsection{Experiment Site Description}
\label{experiment_site_description}

In this study, 12 Holstein Friesian \citep{britannica_holstein} cows were selected on a small-scale dairy farm in the Ropar and Tarn Taran region of Punjab. Our research takes a comprehensive approach by deploying an animal behavior detection system that utilizes a camera installed at an elevated height. Each camera oversees the behavior and movements of the livestock in the area of approx 100m x 80m, as shown in figure \ref{fig:inside_view} and \ref{fig:outside_view}. The herd is provided with a designated area for walking, accessible food and water pots during the experiment, allowing them to stay nourished and hydrated at their convenience. These multi-angle cameras are strategically installed for targeted observation of livestock activities. One camera provides insights into indoor behavior (figure \ref{fig:inside_view}), and the other one captures the outdoor behavior exhibited during outdoor activities (figure \ref{fig:outside_view}). These cameras provide clear and detailed observations even in low-light conditions. The footage captured by the cameras is transmitted and stored for further analysis.

\subsection{Sensor placement in cows} \label{subsubsec2}
\begin{figure}[htbp]
    \centering

    \begin{subfigure}[t]{0.48\textwidth}
        \centering
        \fbox{\includegraphics[width=0.9\textwidth, height=4.1cm]{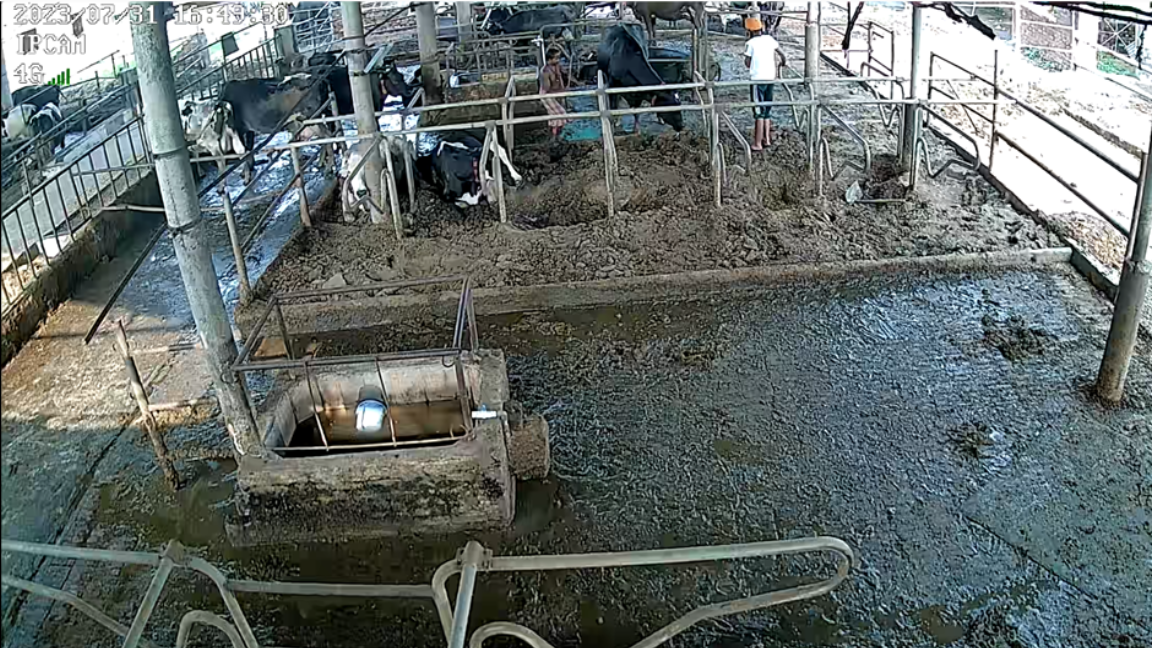}}
        \caption{} \label{fig:inside_view}
    \end{subfigure} \hfill
    \begin{subfigure}[t]{0.48\textwidth}
        \centering
        \fbox{\includegraphics[width=0.9\textwidth, height=4.1cm]{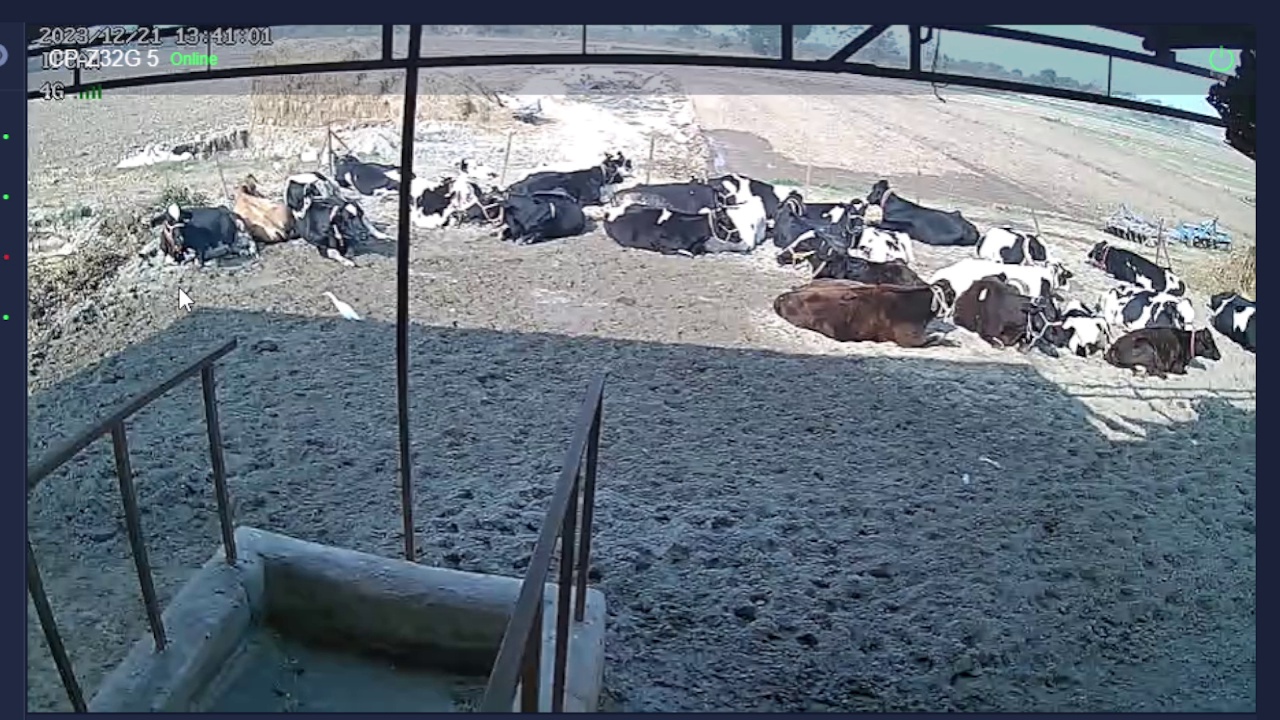}}
        \caption{} \label{fig:outside_view}
    \end{subfigure}

    \vspace{0.5em} 

    \begin{subfigure}[t]{0.48\textwidth}
        \centering
        \fbox{\includegraphics[width=0.9\textwidth, height=4.1cm]{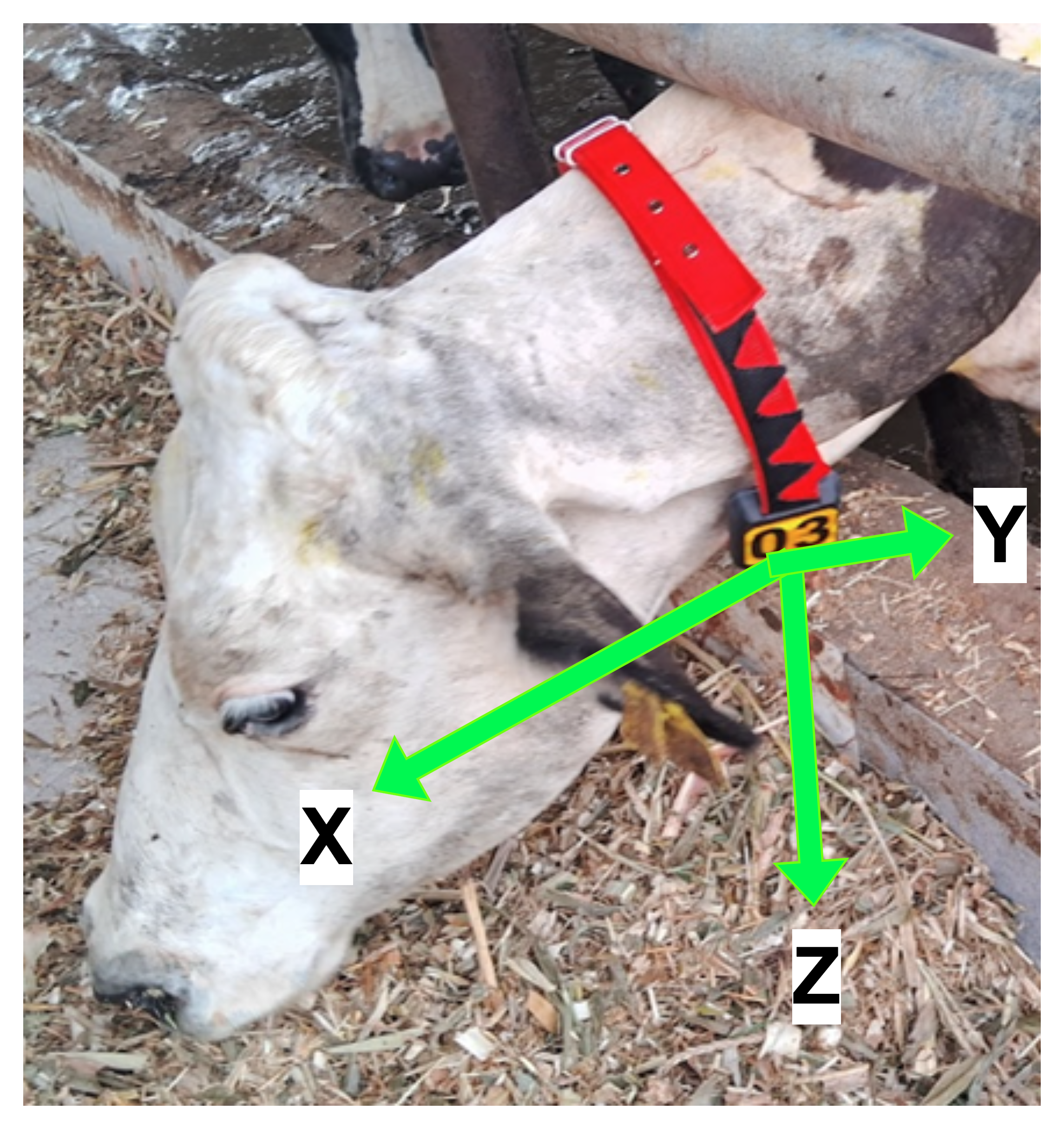}}
        \caption{} \label{fig:cow_with_node}
    \end{subfigure} \hfill
    \begin{subfigure}[t]{0.48\textwidth}
        \centering
        \fbox{\includegraphics[width=0.9\textwidth, height=4.1cm]{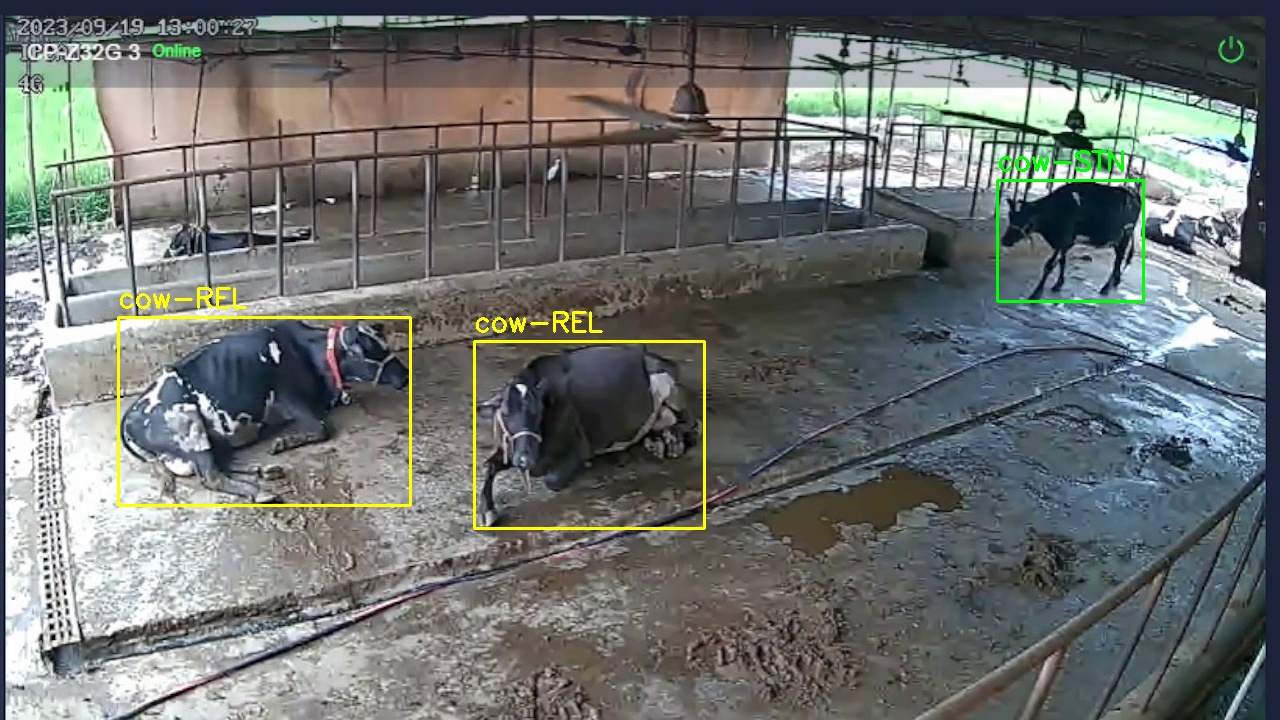}}
        \caption{} \label{fig:Annotations}
    \end{subfigure}

    \caption{\textbf{Farm and Cow's View:}
    (\textbf{\subref{fig:inside_view}}) Inside view of the farm,
    (\textbf{\subref{fig:outside_view}}) Outside view of the farm,
    (\textbf{\subref{fig:cow_with_node}}) Node on the cow's neck,
    (\textbf{\subref{fig:Annotations}}) Sample annotations.}
    \label{fig:farm_and_node}
\end{figure}

The 3-axis accelerometer is placed on the neck of the cows as shown in Figure (\ref{fig:cow_with_node}), and in each pan, 5 to 10 cows are present in such a way that each cow is visible from the camera. Proper visibility is an important factor in ensuring that the annotation is done correctly.  Meanwhile, accelerometer devices transmit data to a 4G-enabled gateway, which processes the data into packets and forwards it to the AWS cloud. A fully automated cloud solution is deployed to store the data and generate inferences from it. In case of bad network connection, the data is stored in the flash memory, which is later uploaded to the cloud. In case of a power outage Uninterruptible Power Supply(UPS) provides power backup for up to 5 hours.

\subsubsection{Data Collection} \label{subsubsec3}

For annotation, cow activities are observed and recorded using a CCTV camera and stored in the local system for later use. The video is recorded from morning to evening and captures approximately 5 hours of footage per day. Each video features embedded timestamps that are analyzed by the annotation team, Annotators trawl through video footage and carefully annotate the raw data into different categories. A person is also assigned to the farm to capture the data manually. The manually annotated data from the on-site observer at the farm and the annotations from the annotation team are compared to ensure consistency and alignment in the annotations.\par


\begin{figure}[htbp]
    \centering
    \begin{subfigure}[t]{0.48\textwidth}
        \centering
        \fbox{\includegraphics[width=0.95\linewidth, height=4.1cm]{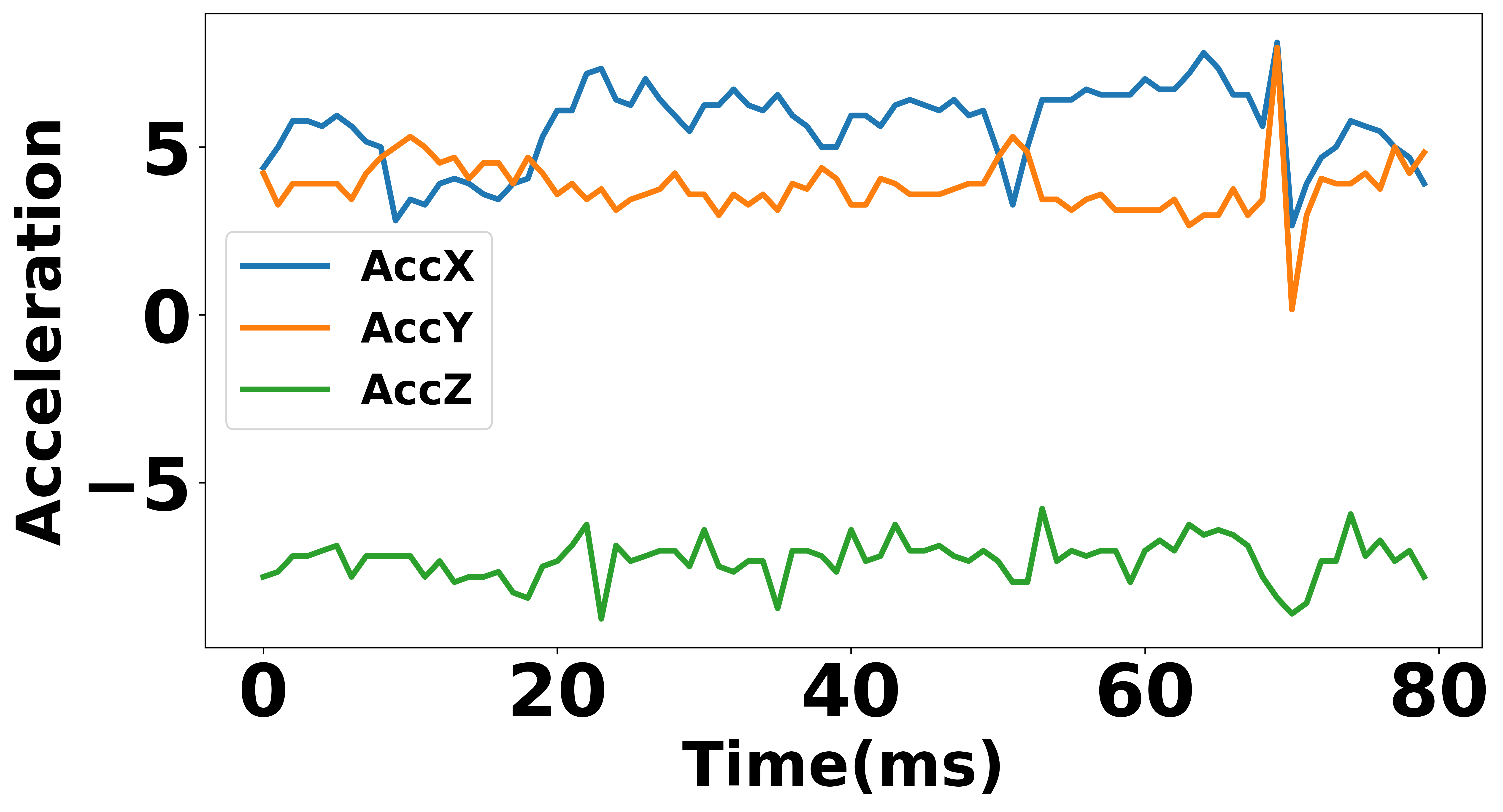}}
        \caption{} \label{fig:STN}
    \end{subfigure} \hfill
    \begin{subfigure}[t]{0.48\textwidth}
        \centering
        \fbox{\includegraphics[width=0.95\linewidth, height=4.1cm]{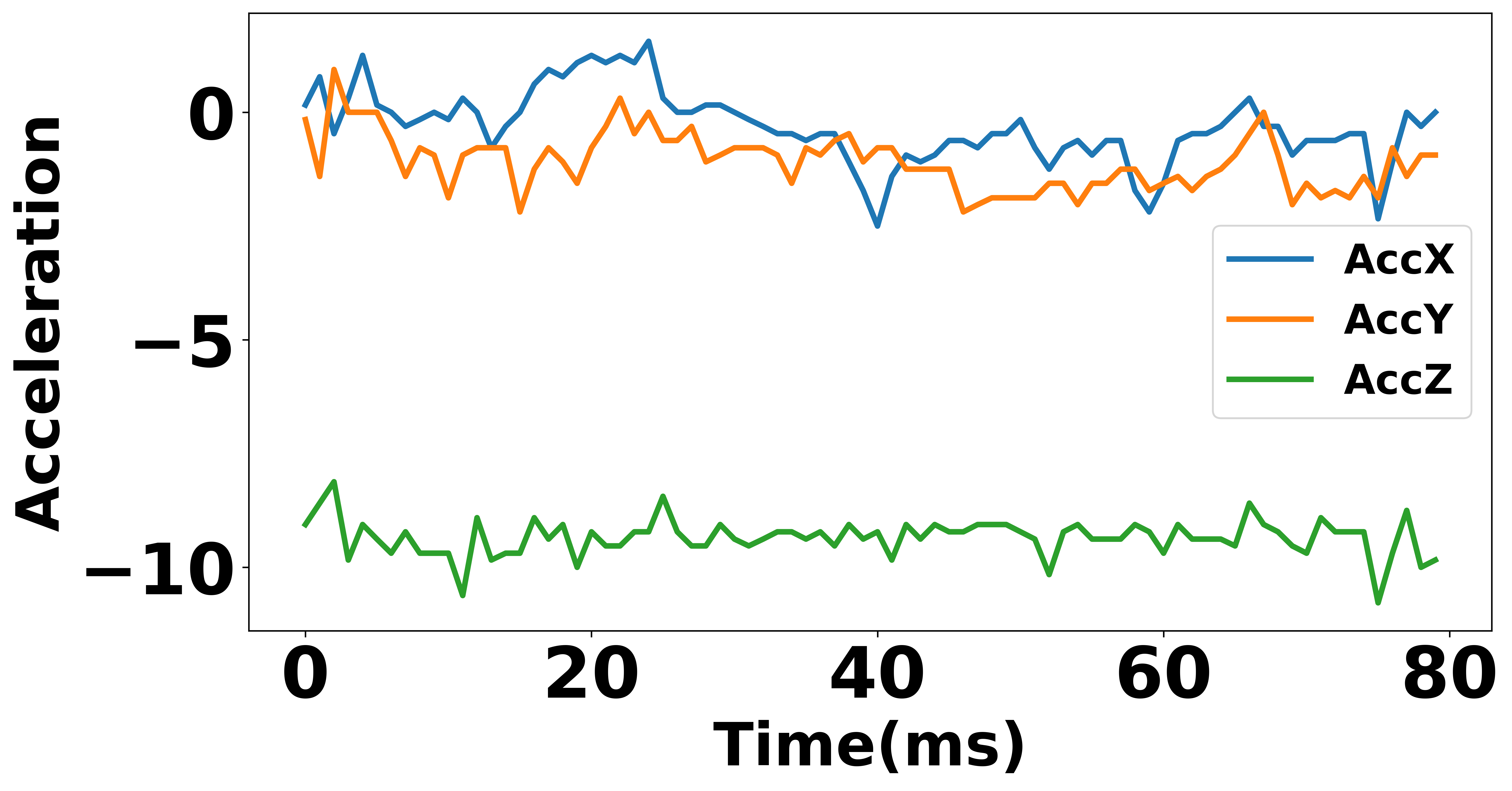}}
        \caption{} \label{fig:REL}
    \end{subfigure}

    \vspace{0.5em} 

    \begin{subfigure}[t]{0.48\textwidth}
        \centering
        \fbox{\includegraphics[width=0.95\linewidth, height=4.1cm]{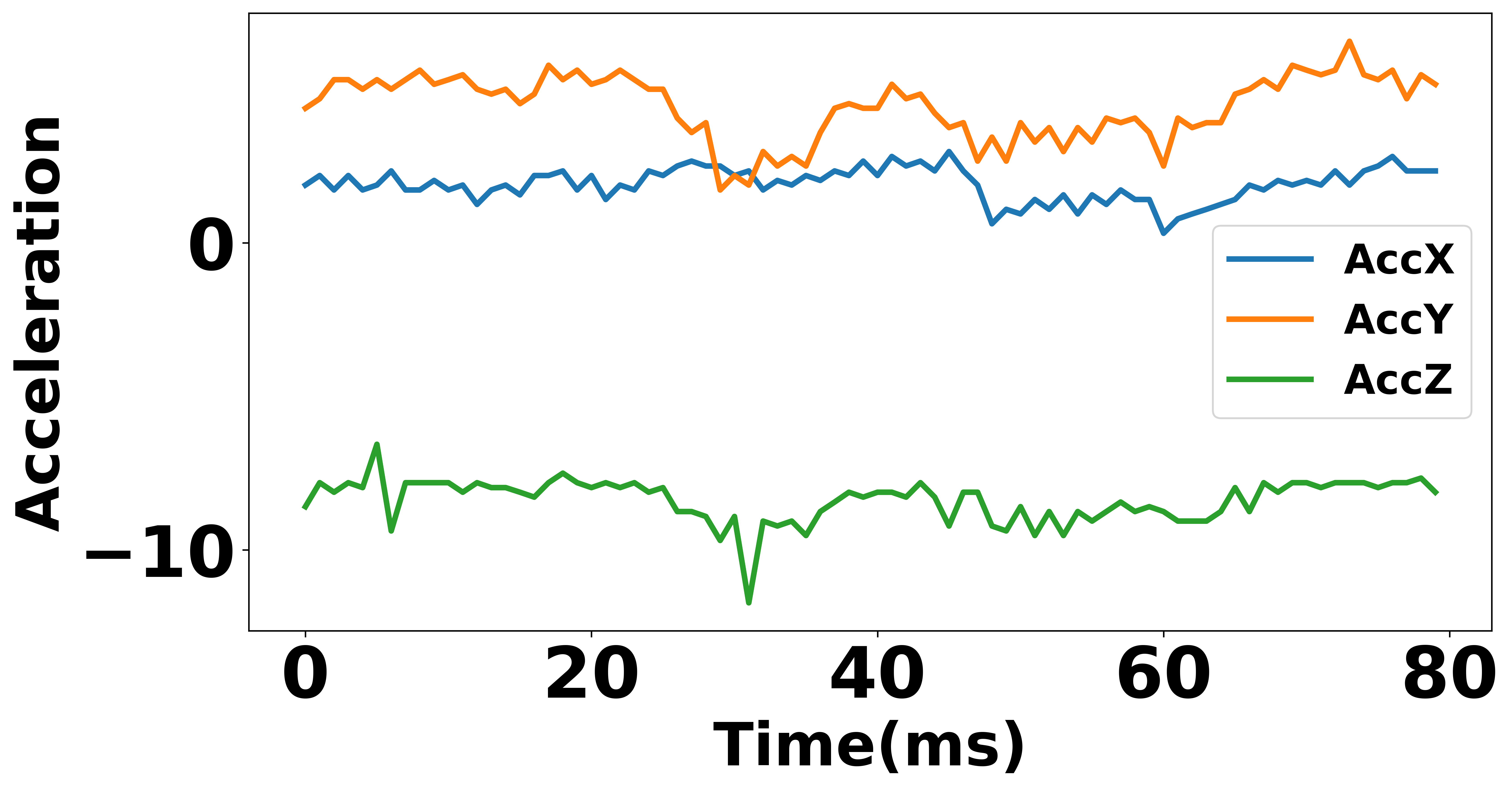}}
        \caption{} \label{fig:RUS}
    \end{subfigure} \hfill
    \begin{subfigure}[t]{0.48\textwidth}
        \centering
        \fbox{\includegraphics[width=0.95\linewidth, height=4.1cm]{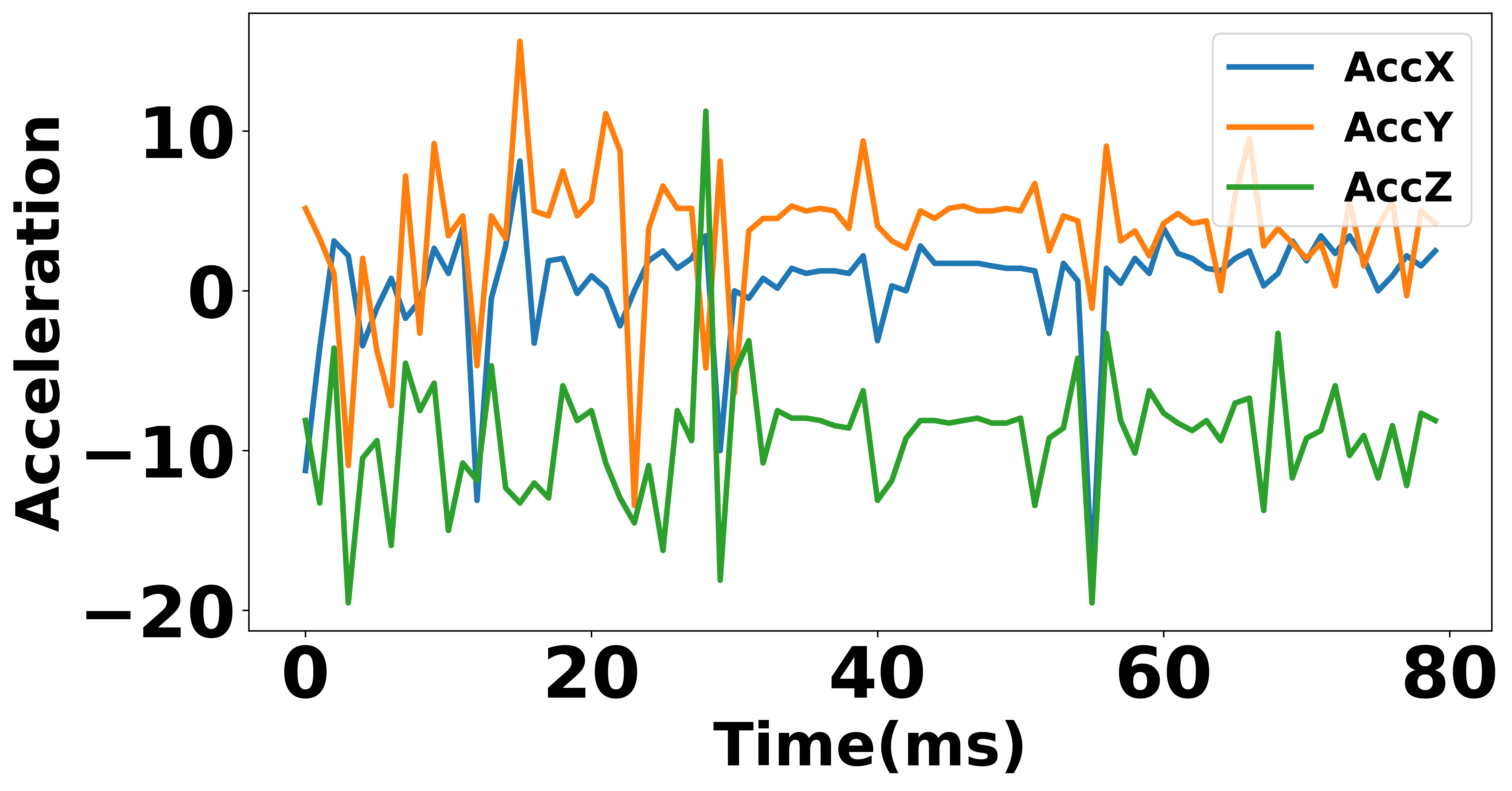}}
        \caption{} \label{fig:ETC}
    \end{subfigure}

    \caption{\textbf{Accelerometer Reading for Different Activities:}
    (\textbf{\subref{fig:STN}}) Standing,
    (\textbf{\subref{fig:REL}}) Resting in Lying,
    (\textbf{\subref{fig:RUS}}) Ruminating,
    (\textbf{\subref{fig:ETC}}) Miscellaneous Activity.}
    \label{fig:accelerometer_reading}
\end{figure}

The dataset, built from the collection of observed behaviors, serves as the cornerstone of our study. The study includes 10 behaviors such as Feeding in Pot (FEP), Attacking (ATT), Licking( LCK), Defecating (DEF), Drinking (DRN), Urinating (URI), Moving (MOV), Resting in a standing position (RES), Resting in a lying position (REL) and  Ruminating in a standing position(RUS) mentioned in table \ref{tab:activity}. This carefully compiled dataset comprises a data file containing four columns, where the first three represent accelerations along the X, Y, and Z axes, respectively, and the fourth column records their corresponding behavioral class.
Figure \ref{fig:accelerometer_reading} illustrates the typical recorded acceleration signals for distinct behaviors in the cow dataset. It is evident from the figure that the acceleration patterns of various behavioral categories display unique attributes, providing valuable insights into their corresponding movements and activities.

\begin{table}[ht]
\centering
\caption{Behavior definitions}
\label{tab:activity}
\renewcommand{\arraystretch}{1.2}
\begin{tabular}{c   c   c}
\hline
\rule{0pt}{2.5ex}
\textbf{Abbreviations} & \textbf{Datapoint Count}  & \textbf{Description} \\ 
\hline
RES & 226561 & Resting in standing position \\
RUS & 195600 & Ruminating in standing position \\
REL & 130960 & Resting in lying position \\
FEP & 112079 & Feeding in Pot \\
MOV & 16320 & Moving \\
LCK & 8240 & Licking \\
ATT & 1280 & Attacking \\
DEF & 960 & Defecating \\
DRN & 880 & Drinking \\
URI & 640 & Urinating \\
\hline
\end{tabular}
\end{table}

\subsection{Data Annotation}
\begin{figure}[htbp]
    \centering
    \fbox{
        \includegraphics[width=\linewidth]{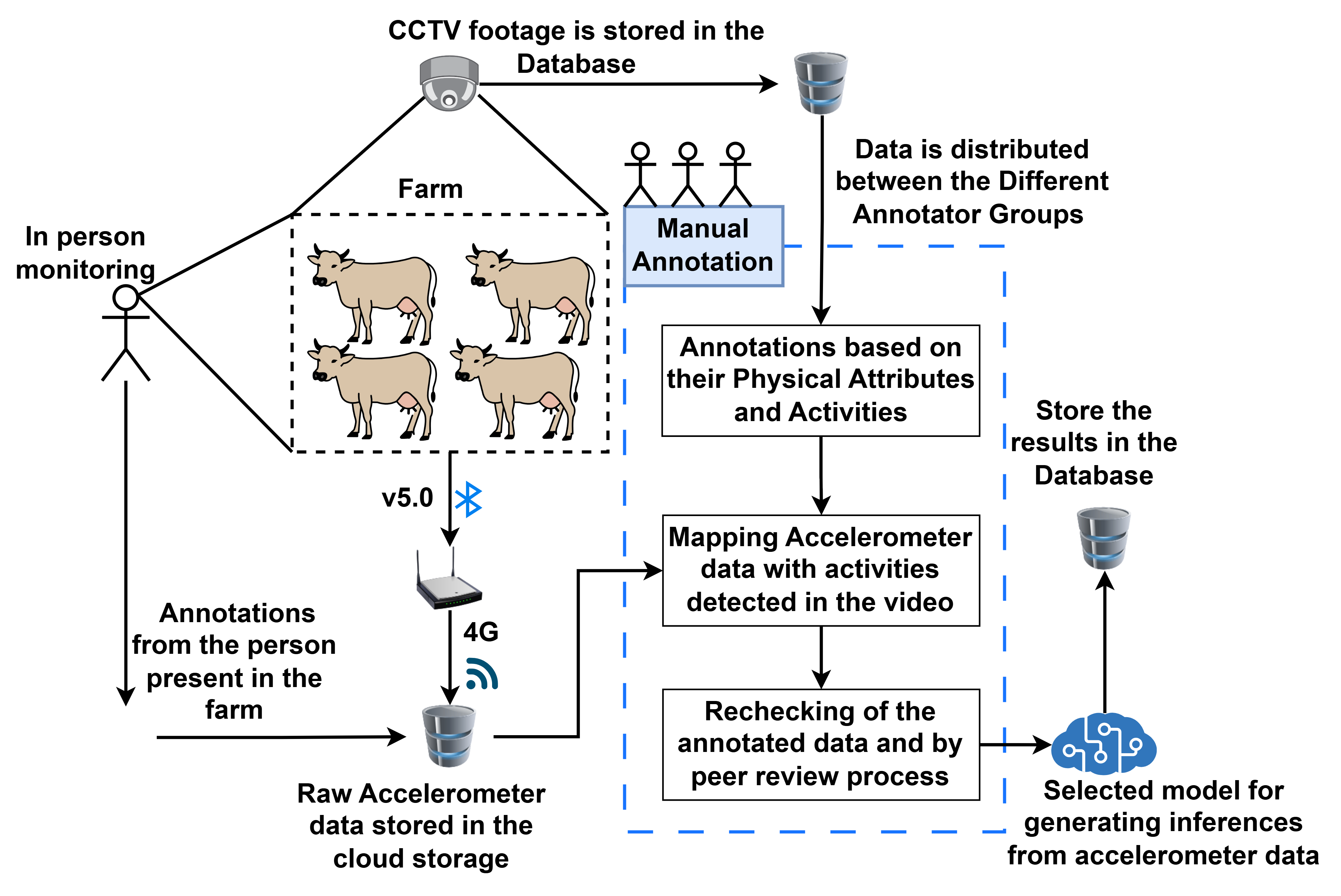}
    }
    \caption{Data collection and annotation pipeline.}
    \label{fig:mthodology}
\end{figure}

Acquisition and annotation of the data is the most resource-intensive work in terms of both time and cognitive bandwidth consumption. This involves the following systematic steps to ensure accuracy and efficiency in capturing behavioral patterns as displayed in figure \ref{fig:mthodology}

\noindent\textbf{Step 1:} Cows are placed in dedicated pans (each pan contains 5 - 7 cows), ensuring they are easily visible by the on-site observer and in the video footage.

\noindent \textbf{Step 2:} Detailed video and images of the targeted cows are taken. The images and videos are provided to annotators for detecting and identifying cows along with their corresponding node IDs.

\noindent \textbf{Step 3:} Cameras are positioned at two different angles to comprehensively monitor the cows' activities throughout the day, covering both indoor and outdoor environments. The observer present at the farm manually notes down the cows' activities. These manually noted activities serve as the ground truth for video annotations.

\noindent \textbf{Step 4:} Accelerometer readings are  transmitted to the gateway device via Bluetooth connection(BLE 4.0). The gateway device forwards the data to the cloud storage using a 4G connection.

\noindent \textbf{Step 5:} Video data from camera devices are also captured, stored, and distributed for annotations.

\noindent \textbf{Step 6:} Relevant video data is extracted from the video files and forwarded to the annotators for manual review to identify activity-specific data points.

\noindent \textbf{Step 7:} Extracted activities are mapped with the corresponding timestamps.

\noindent \textbf{Step 8:} Once the activities and timestamps are mapped, the data goes through a peer review process as a secondary validation step. During this step, video annotations are cross-checked against in-person annotations, where the latter is considered the ground truth.

\noindent \textbf{Step 9:} The data is sent to the subsequent stages of the pipeline, such as feature extraction, selection, and model training.

\subsection{Data Preprocessing and Feature Extraction}\label{Data_Preprocessing}
Two crucial aspects are emphasized during the pre-processing step: Data cleaning and feature extraction, each playing a pivotal role in preparing the data for subsequent analysis. The algorithm \ref{algo:feat_ext_algo} extracts distinctive features from tri-axial accelerometer data. The first category, known as instantaneous features, provides a snapshot of activities at specific moments, which includes raw acceleration data of the X-axis, Y-axis, and Z-axis, signal magnitude area, vector magnitude, movement variation, energy, and entropy. These features capture the intensity of activities within a particular second. In addition, angle characteristics (roll, yaw, pitch) were also incorporated to provide insights on device orientation.\par 
The second category involves statistical features calculated over the sliding windows, which include mean, standard deviation, sum, variance, mean absolute deviation, median, minimum, maximum, quartiles, kurtosis, and skewness. The large time-series data is segmented into smaller frames to reduce computational resource consumption. This method captures temporal patterns over time and is particularly useful for understanding trends. The parameters defining the sliding window, such as size, sampling rate, and step size, are flexible and can be adjusted based on individual preference. To enhance the depth of our analysis, we introduced Discrete Wavelet Transform (DWT) features. These features, including mean, variance, and energy, were computed for both the approximate and detail coefficients. DWT offers multi-resolution analysis, allowing us to capture variations in the accelerometer data across different frequency bands, thus expanding the scope of our feature set and providing valuable insights into the frequency characteristics of the recorded activities.\par

\subsubsection{Lag Features Generation}
To capture the temporal nature of the data, we have implemented the lagged data(up to lag 5) features. To generate lag features, we use the previous data points as a feature. Accelerometer data is a time series in nature, which is collected and organized based on the timestamp sequences. The previous 5 data points were used as lag features for training the models. For example, if we have a feature $\text{AccX\_median}$ with 4.5 as the current value, after lag implementation, we will have 5 more attributes, \textit{Acc\_X\_lag\_1}, \textit{Acc\_X\_lag\_2}, \textit{Acc\_X\_lag\_3}, \textit{Acc\_X\_lag\_4}, \textit{Acc\_X\_lag\_5}. Here, \textit{lag\_1}  represents the immediate previous data point, while \textit{lag\_5}  corresponds to the value received five data points prior. First, we compress a minute's worth of data into a single data point while extracting base features(converting 80 data points to 1), then incorporating lag features to maintain the full temporal nature of the data. Lag features are created by shifting the values of a variable by a targeted number of timesteps.

\begin{algorithm}
\caption{Feature Extraction for Time-Series Data}

\begin{algorithmic}[1]

\Require $dataframe$, $step\_length$, $window\_length$
\Ensure Extract features from DataFrame

\State \textbf{Initialize} $features\_df = pandas.DataFrame$

\State \textbf{Set} $deviceID, labels \gets dataframe[\text{deviceId}, \text{Label}]$ 

\State $range\_end \gets \text{length of }dataframe - window\_length$

\State \textbf{Define} $features\_list$ (mean, kurtosis, skewness, quantile, Energy, Entropy, etc)

\For{$i = 0$ to $range\_end$ with step $step\_length$}
    \State $timestamp\_max \gets \text{max}(dataframe[Timestamp][i : i + window\_length])$
    \State $data\_window \gets [[\text{AccX}, \text{AccY}, \text{AccZ}]][i : i + window\_length]$
    
    \State $window\_features \gets [\text{Roll}, \text{Pitch}]$
    \State \textbf{Apply} $\text{Savitzky-Golay Filter on AccX, AccY and AccZ}$
    \State \textbf{Perform} $\text{Wavelet Decomposition:}$
    \State $wavelet\_coefficients \gets wavelet\_decomposition([AccX, AccY, AccZ], level = 3)$
    \State \textbf{Extract} $\text{Statistical Features such as mean, variance, etc from } wavelet\_coefficients $
    \State $\textbf{Calculate } statistical\_features \gets \text{Statistical features(eg. kurtosis, quantiles, etc)}$
    \State \textbf{Concat different features: } $combined\_features \gets \text{raw\_data, wavelet\_features, statistical\_features}$
\EndFor
\State \textbf{Generate LAG Features:} \text{Add lag Features till 5 previous datapoints.}
\State \textbf{Return:} $combined\_dataframe \gets [combined\_features, lagged\_features]$

\end{algorithmic}

\label{algo:feat_ext_algo}
\end{algorithm}

\subsubsection{Dataset Preparation}
 This is a vital step and involves various processes to ensure the accuracy and consistency of data by removing duplicates, handling missing values, and managing outliers and inconsistencies from data sources. We use the python \textit{pandas} \citep{reback2020pandas} library to clean and pre-process the data. The cleaning process, along with the unequal distribution of data, poses challenges in achieving accurate classification results, especially for the less-represented classes. To address this imbalance, we tested several strategies. Some imbalanced data prevention training methodologies, such as minority undersampling and Synthetic Minority Over-sampling Technique(SMOTE) \citep{chawla2002smote}. However, oversampling did not yield desirable results. Finally, we opted to merge minority classes to create a miscellaneous class, which is denoted as ETC as shown in table \ref{tab:new_classes}. ETC label combines activities such as Feeding in Pot ({FEP), Attacking (ATT), Licking (LCK), Defecating (DEF), Urinating (URI), and Drinking (DRN). Additionally, the STN label is introduced to specifically identify Standing Activity, which includes Resting in a standing position (RES) and Moving (MOV). This refined labeling method enhances the model's ability to discern and categorize distinct cow behaviors during real-time analysis.

\begin{table}[ht]
\centering
\caption{Combined New Classes}
\label{tab:behaviors}
\renewcommand{\arraystretch}{1.2}
\begin{tabular}{c   c   c}
\hline
\rule{0pt}{2.5ex}
\textbf{New class} & \textbf{Datapoints percentage} & \textbf{Old classes} \\  
\hline
STN & 34.98\% & RES and MOV \\
RUS & 28.21\% & RUS  \\
REL & 18.89\% & REL \\
ETC & 17.9\% & FEP, LCK, ATT, DEF, DRN and URI \\
\hline
\label{tab:new_classes}
\end{tabular}
\end{table}

\subsection{Model Development  }\label{sec3}
\subsubsection{Machine Learning Model Development: }
In the deployment of the ML model for cow activity behavior classification, the derived feature file from 3-axial accelerometer data serves as the input for the model. We trained the model using various hyperparameter configurations across different models. Hyperparameter tuning was performed using a grid search. Table \ref{tab:behaviors} outlines the hyperparameters considered and the best values obtained. The optimized hyperparameters were applied during the training phase for each model. Techniques such as under-sampling and oversampling were deemed unsuitable for our use case, as minority under-sampling reduces the overall dataset size, while the significant imbalance between majority and minority classes would undermine the extensive data collection efforts. It was observed that oversampling techniques such as SMOTE add synthetic data points to the dataset, which produce relatively poor overall results. So, we have combined the minority class to create a relatively balanced dataset, which has performed better than any other approach.

\begin{table}[htbp]
    \centering
    \caption{Best Hyperparameters for Each Model Configuration}
    \label{tab:model_hyperparams}
    \begin{tabular}{l c c p{8cm}}
        \hline
        \textbf{Model} & \textbf{Window} & \textbf{Step} & \textbf{Parameters} \\
        \hline
        K-Nearest Neighbors & 156 & 39 & n\_neighbors: 3, p: 1, weights: distance \\
        K-Nearest Neighbors & 316 & 79 & n\_neighbors: 3, p: 1, weights: distance \\
        LightGBM & 316 & 79 & learning\_rate: 0.1, max\_depth: 7, n\_estimators: 200, reg\_alpha: 0, reg\_lambda: 0.1 \\
        LightGBM & 156 & 39 & learning\_rate: 0.1, max\_depth: 7, n\_estimators: 200, reg\_alpha: 0.1, reg\_lambda: 0.01 \\
        Gradient Boosting & 316 & 79 & learning\_rate: 0.1, max\_depth: 7, n\_estimators: 200 \\
        XGBoost & 156 & 39 & learning\_rate: 0.1, max\_depth: 7, n\_estimators: 200, reg\_alpha: 0.1, reg\_lambda: 0.01 \\
        XGBoost & 316 & 79 & learning\_rate: 0.1, max\_depth: 7, n\_estimators: 200, reg\_alpha: 0.1, reg\_lambda: 0.01 \\
        Gradient Boosting & 156 & 39 & learning\_rate: 0.1, max\_depth: 7, n\_estimators: 200 \\
        Random Forest & 316 & 79 & max\_depth: None, min\_samples\_split: 2, n\_estimators: 200 \\
        Random Forest & 156 & 39 & max\_depth: 30, min\_samples\_split: 2, n\_estimators: 100 \\
        \hline
    \end{tabular}
\end{table}

\subsubsection{Hyperparameter Searching and Model Selection: }
The process of hyperparameter searching and model selection involves the following systematic steps to ensure optimal model performance.

\noindent \textbf{Step 1:} A List of hyperparameters is selected for all models. This includes the algorithm's standard parameters and also the window and step length. As we have shown in the table \ref{tab:model_hyperparams}.

\noindent \textbf{Step 2:} The dataset is divided into training (60\%), validation (20\%), and testing (20\%) subsets, ensuring that the class ratios are preserved during the split.

\noindent \textbf{Step 3:} To prevent data leakage during searching testing set was segregated and utilized exclusively for evaluation purposes.

\noindent \textbf{Step 4:} 5-fold grid search \citep{scikit-learn_grid_search} was used for hyperparameter searching. Once the search is completed, we run an evaluation where we consider the accuracy, precision, recall, and F1 score.

\noindent \textbf{Step 5:} Based on the search, we observed that approaches such as Random Forest, k-Nearest Neighbors, and Gradient Boosting algorithms performed better. k-Nearest Neighbors seems to be the better with training accuracy(\textbf{0.926 $\pm$ 0.0053})  and testing accuracy(\textbf{0.939}).

\begin{figure}[htbp]
    \centering
    \begin{subfigure}[t]{0.48\textwidth}
        \centering
        \fbox{\includegraphics[width=0.95\linewidth, height=5cm]{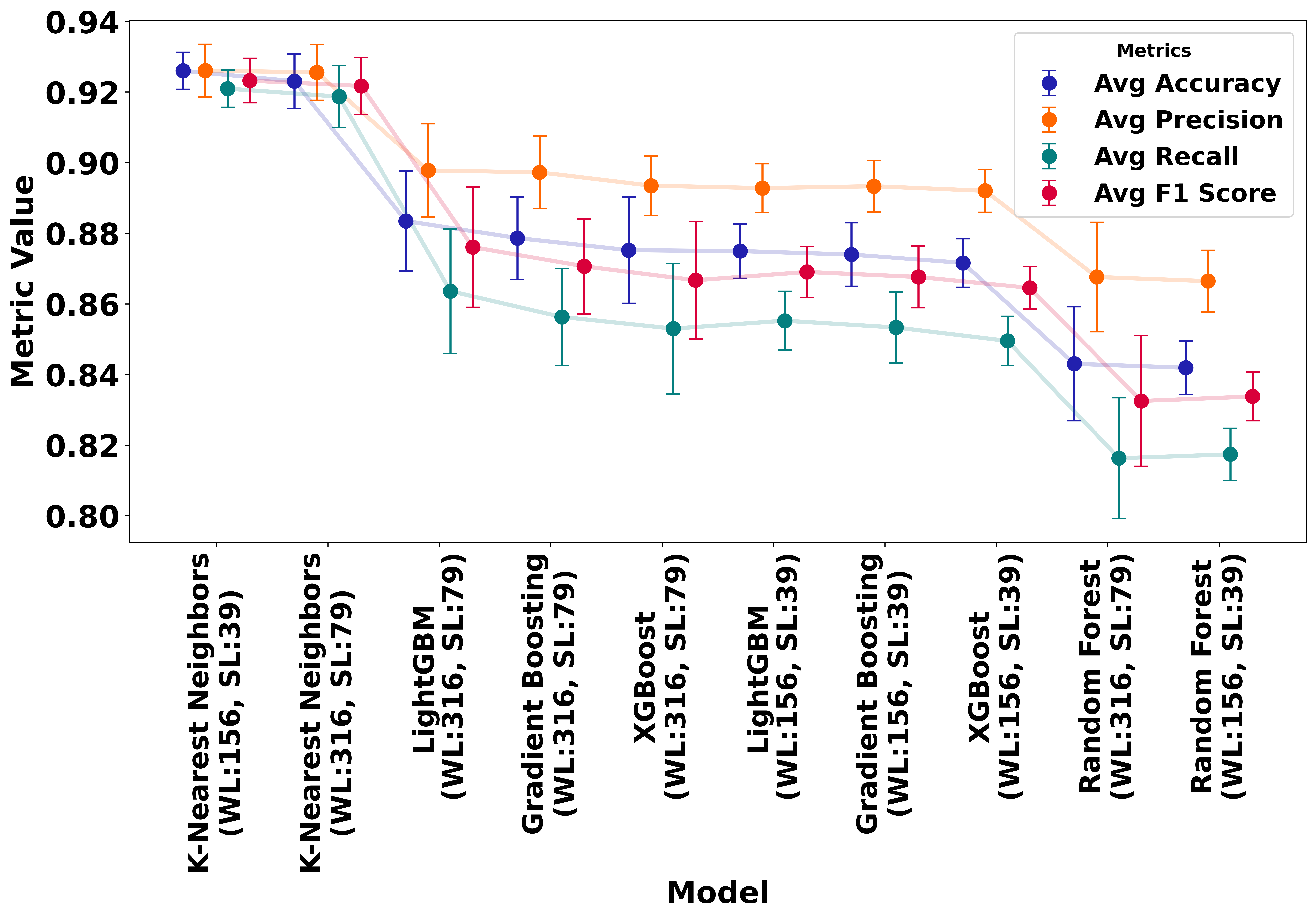}}
        \caption{} \label{fig:training Performance comparision}
    \end{subfigure} \hfill
    \begin{subfigure}[t]{0.48\textwidth}
        \centering
        \fbox{\includegraphics[width=0.95\linewidth, height=5cm]{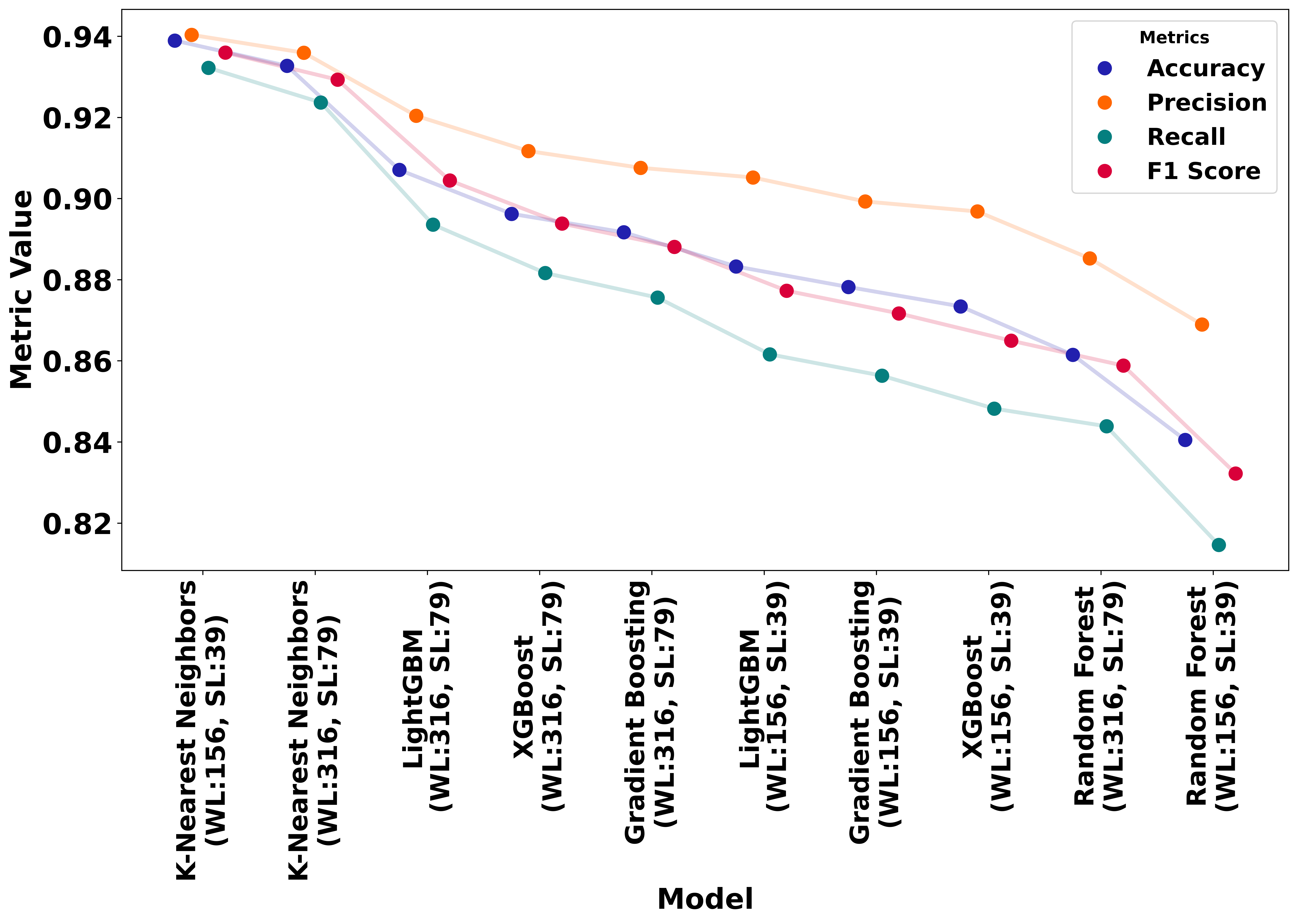}}
        \caption{} \label{fig:testing Performance comparision}
    \end{subfigure}

    \vspace{0.5em} 

    \begin{subfigure}[t]{0.48\textwidth}
        \centering
        \fbox{\includegraphics[width=0.95\linewidth, height=5cm]{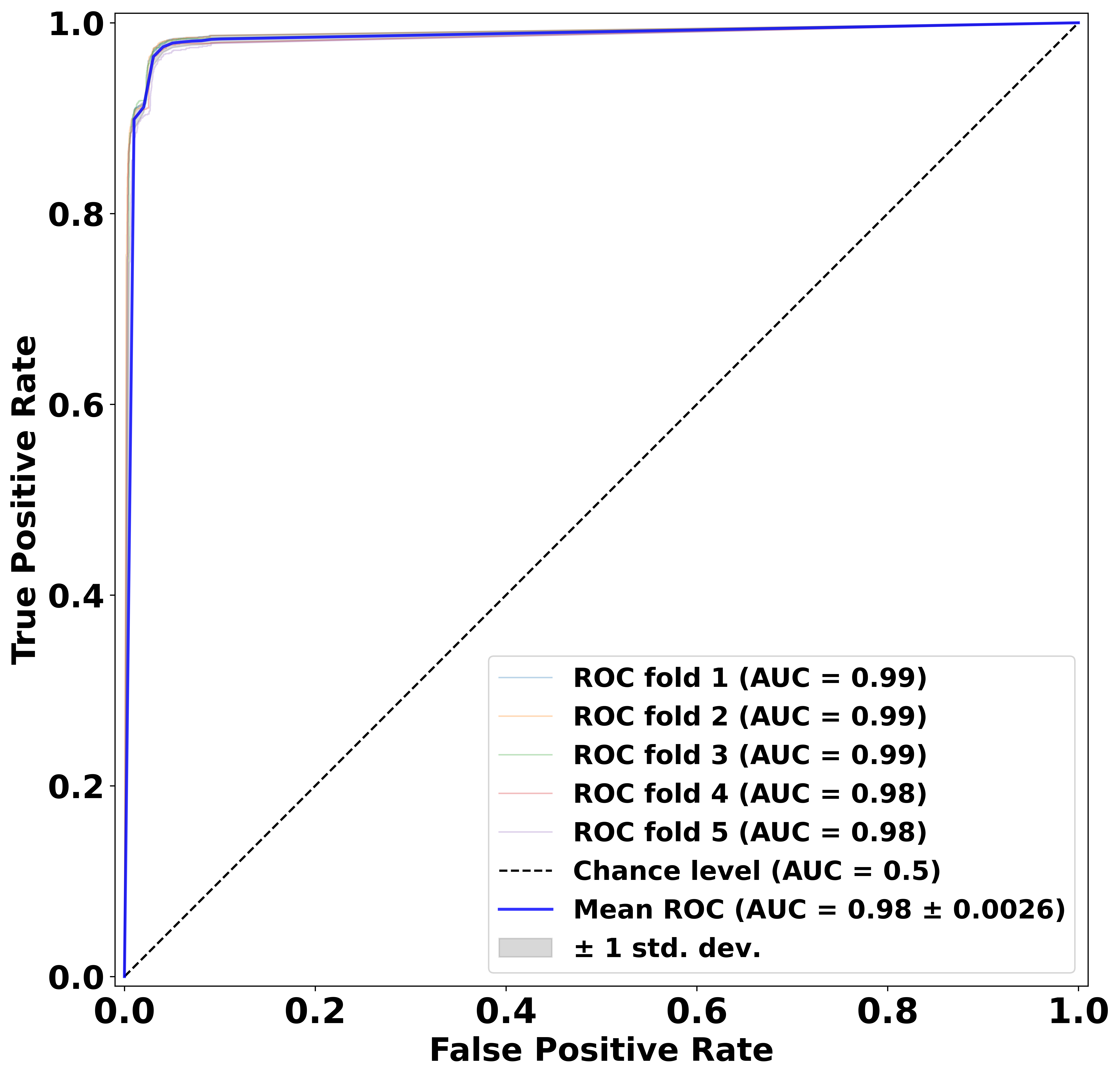}}
        \caption{} \label{fig:AUC_train}
    \end{subfigure} \hfill
    \begin{subfigure}[t]{0.48\textwidth}
        \centering
        \fbox{\includegraphics[width=0.95\linewidth, height=5cm]{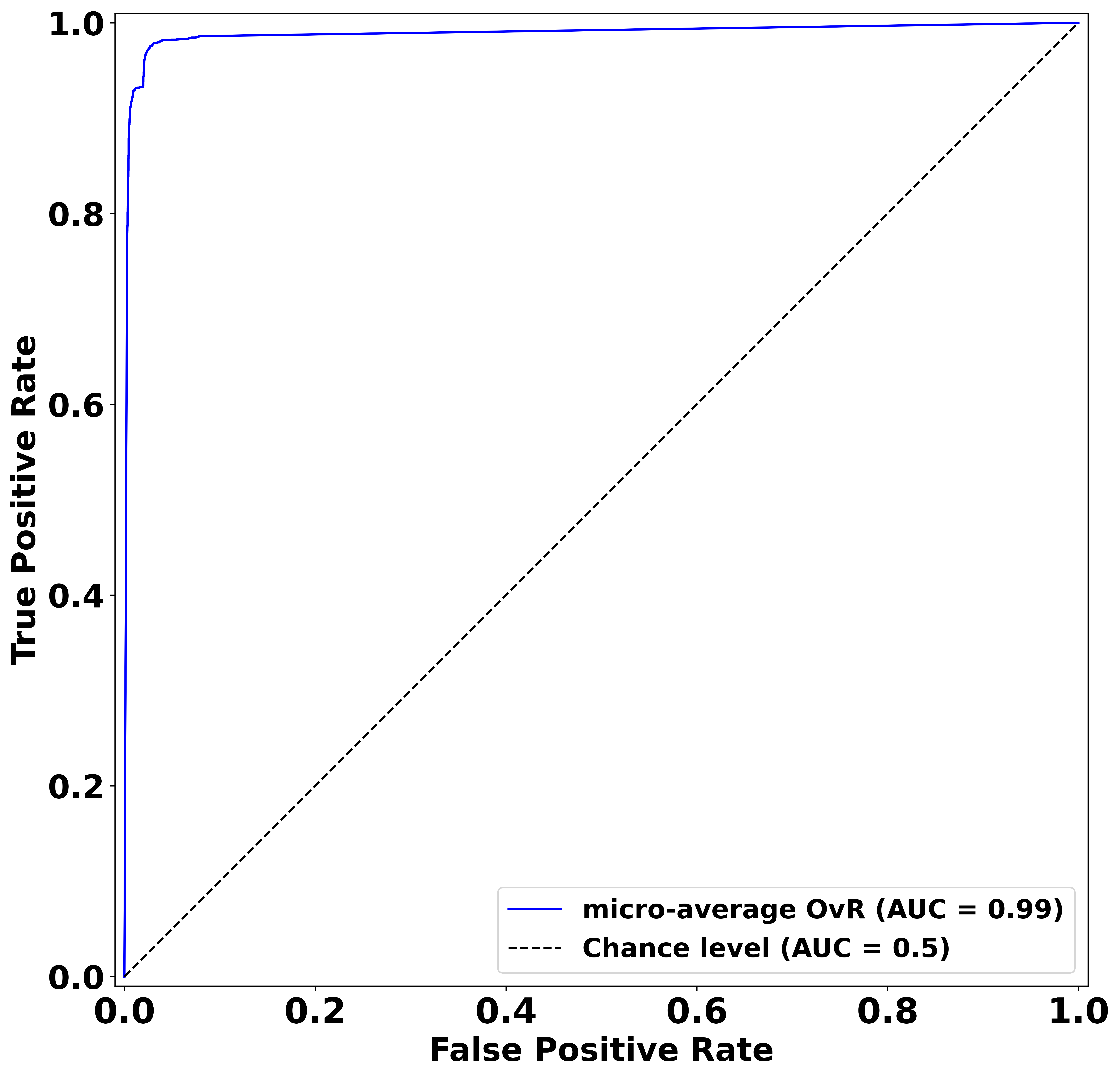}}
        \caption{} \label{fig:AUC_test}
    \end{subfigure}
    
    \vspace{0.5em} 

    \caption{\textbf{Accelerometer Reading for Different Activities:}
    (\textbf{\subref{fig:training Performance comparision}}) Model Performance Analysis for Training Data.
    (\textbf{\subref{fig:testing Performance comparision}}) Model Performance Analysis for Training Data.
    (\textbf{\subref{fig:AUC_train}}) ROC-AUC Curve(0.98 $\pm$ 0.0026) for Training Set for best performing KNN.
    (\textbf{\subref{fig:AUC_test}}) ROC-AUC Curve(0.99) for Testing Set for best performing KNN.}
    \label{fig:Model performance comparision}
\end{figure}

\begin{table}[htbp]
    \centering
    \caption{Model Performance Metrics with Standard Deviation on Training set}
    \label{table:model_performance_train}
    \begin{tabular}{l c c c c}
        \hline
        \textbf{Model (Window/Step)} & \textbf{Accuracy} & \textbf{Precision} & \textbf{Recall} & \textbf{F1 Score} \\
        \hline
        \textbf{KNN (156/39)} & \textbf{0.926 $\pm$ 0.0053} & \textbf{0.926 $\pm$ 0.0075} & \textbf{0.921 $\pm$ 0.0052} & \textbf{0.923 $\pm$ 0.0063} \\
        KNN (316/79) & 0.923 $\pm$ 0.0077 & 0.926 $\pm$ 0.0079 & 0.919 $\pm$ 0.0088 & 0.922 $\pm$ 0.0081 \\
        LightGBM (316/79) & 0.883 $\pm$ 0.0142 & 0.898 $\pm$ 0.0132 & 0.864 $\pm$ 0.0176 & 0.876 $\pm$ 0.0170 \\
        Gradient Boosting (316/79) & 0.879 $\pm$ 0.0117 & 0.897 $\pm$ 0.0103 & 0.856 $\pm$ 0.0137 & 0.871 $\pm$ 0.0134 \\
        XGBoost (316/79) & 0.875 $\pm$ 0.0150 & 0.893 $\pm$ 0.0084 & 0.853 $\pm$ 0.0185 & 0.867 $\pm$ 0.0167 \\
        LightGBM (156/39) & 0.875 $\pm$ 0.0077 & 0.893 $\pm$ 0.0069 & 0.855 $\pm$ 0.0083 & 0.869 $\pm$ 0.0073 \\
        Gradient Boosting (156/39) & 0.874 $\pm$ 0.0090 & 0.893 $\pm$ 0.0073 & 0.853 $\pm$ 0.0100 & 0.868 $\pm$ 0.0087 \\
        XGBoost (156/39) & 0.872 $\pm$ 0.0068 & 0.892 $\pm$ 0.0061 & 0.850 $\pm$ 0.0070 & 0.865 $\pm$ 0.0060 \\
        Random Forest (316/79) & 0.843 $\pm$ 0.0162 & 0.868 $\pm$ 0.0155 & 0.816 $\pm$ 0.0171 & 0.832 $\pm$ 0.0185 \\
        Random Forest (156/39) & 0.842 $\pm$ 0.0076 & 0.866 $\pm$ 0.0087 & 0.817 $\pm$ 0.0074 & 0.834 $\pm$ 0.0069 \\
        \hline
    \end{tabular}
\end{table}

\begin{table}[htbp]
    \centering
    \caption{Model Performance Comparison with Window and Step Sizes on Test set}
    \label{tab:model_comparison_test}
    \begin{tabular}{l c c c c}
        \hline
        \textbf{Model (Window/Step)} & \textbf{Accuracy} & \textbf{Precision} & \textbf{Recall} & \textbf{F1 Score} \\
        \hline
        \textbf{K-Nearest Neighbors (156/39)} & \textbf{0.9389} & \textbf{0.9404} & \textbf{0.9322} & \textbf{0.9360} \\
        K-Nearest Neighbors (316/79) & 0.9327 & 0.9359 & 0.9237 & 0.9293 \\
        LightGBM (316/79) & 0.9071 & 0.9204 & 0.8936 & 0.9045 \\
        XGBoost (316/79) & 0.8962 & 0.9117 & 0.8816 & 0.8938 \\
        Gradient Boosting (316/79) & 0.8917 & 0.9076 & 0.8756 & 0.8881 \\
        LightGBM (156/39) & 0.8833 & 0.9052 & 0.8616 & 0.8773 \\
        Gradient Boosting (156/39) & 0.8782 & 0.8993 & 0.8563 & 0.8717 \\
        XGBoost (156/39) & 0.8734 & 0.8968 & 0.8482 & 0.8650 \\
        Random Forest (316/79) & 0.8615 & 0.8852 & 0.8439 & 0.8588 \\
        Random Forest (156/39) & 0.8405 & 0.8690 & 0.8146 & 0.8322 \\
        \hline
    \end{tabular}
\end{table}

\subsubsection{ML model's performance comparison: }
Several models, such as KNN, XGBoost, Gradient Boost, Random Forest, Logistic Regression, decision tree, and  Naive Bayes, are used for model selection. Table \ref{table:model_performance_train}. As observed from the Table, the KNN model with a window length of 156 and a step length of 39 outperforms other models on all the metrics. Figures (\ref{fig:training Performance comparision}) and (\ref{fig:testing Performance comparision}) show the performance comparison of the top 10 models, which corresponds to table \ref{table:model_performance_train}. Figure \ref{fig:AUC_train} and \ref{fig:AUC_test} show the model's AUC scores, and Figure \ref{fig:cfm} shows the confusion matrix for the testing set. Table \ref{tab:model_comparison_test} compares the model performance with the window size and step size in the test set.

\begin{figure}[htbp]
    \centering
    \begin{subfigure}[t]{0.48\textwidth}
        \centering
        \fbox{\includegraphics[width=0.95\linewidth, height=5cm]{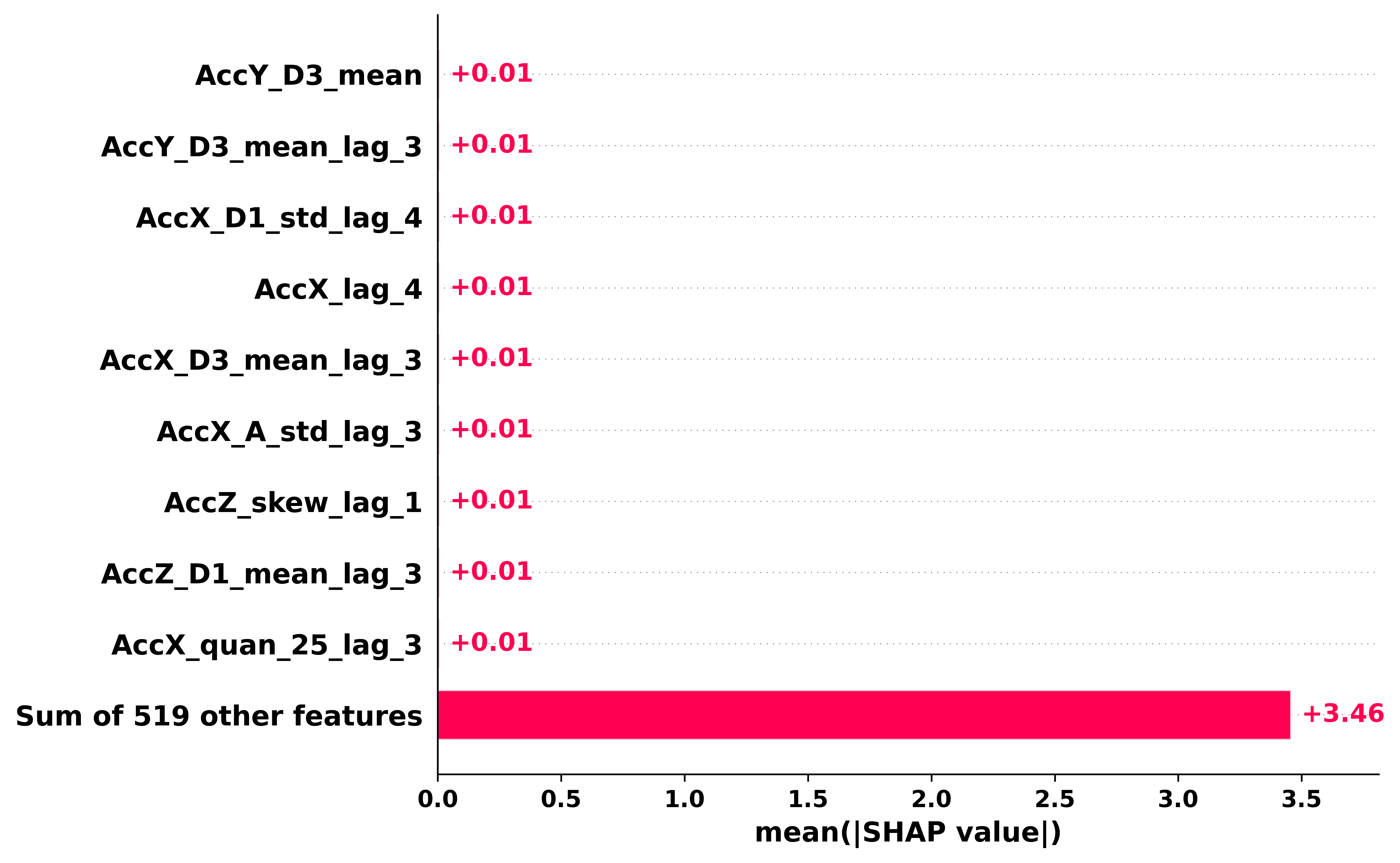}}
        \caption{} \label{fig:shap_bar_STN}
    \end{subfigure}
    \hfill
    \begin{subfigure}[t]{0.48\textwidth}
        \centering
        \fbox{\includegraphics[width=0.95\linewidth, height=5cm]{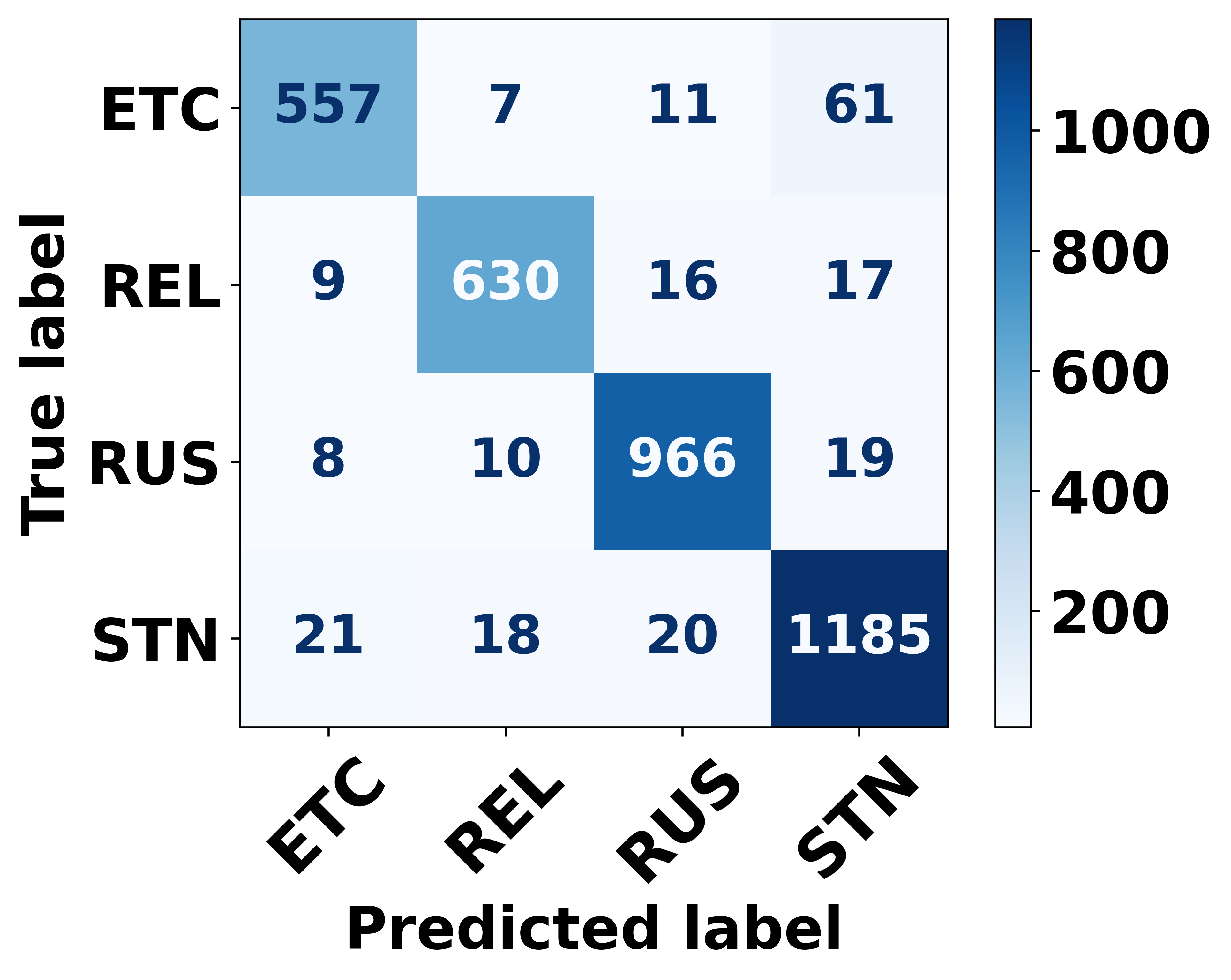}}
        \caption{} \label{fig:cfm}
    \end{subfigure}
    \hfill
    \begin{subfigure}[t]{0.48\textwidth}
        \centering
        \fbox{\includegraphics[width=0.95\linewidth, height=5cm]{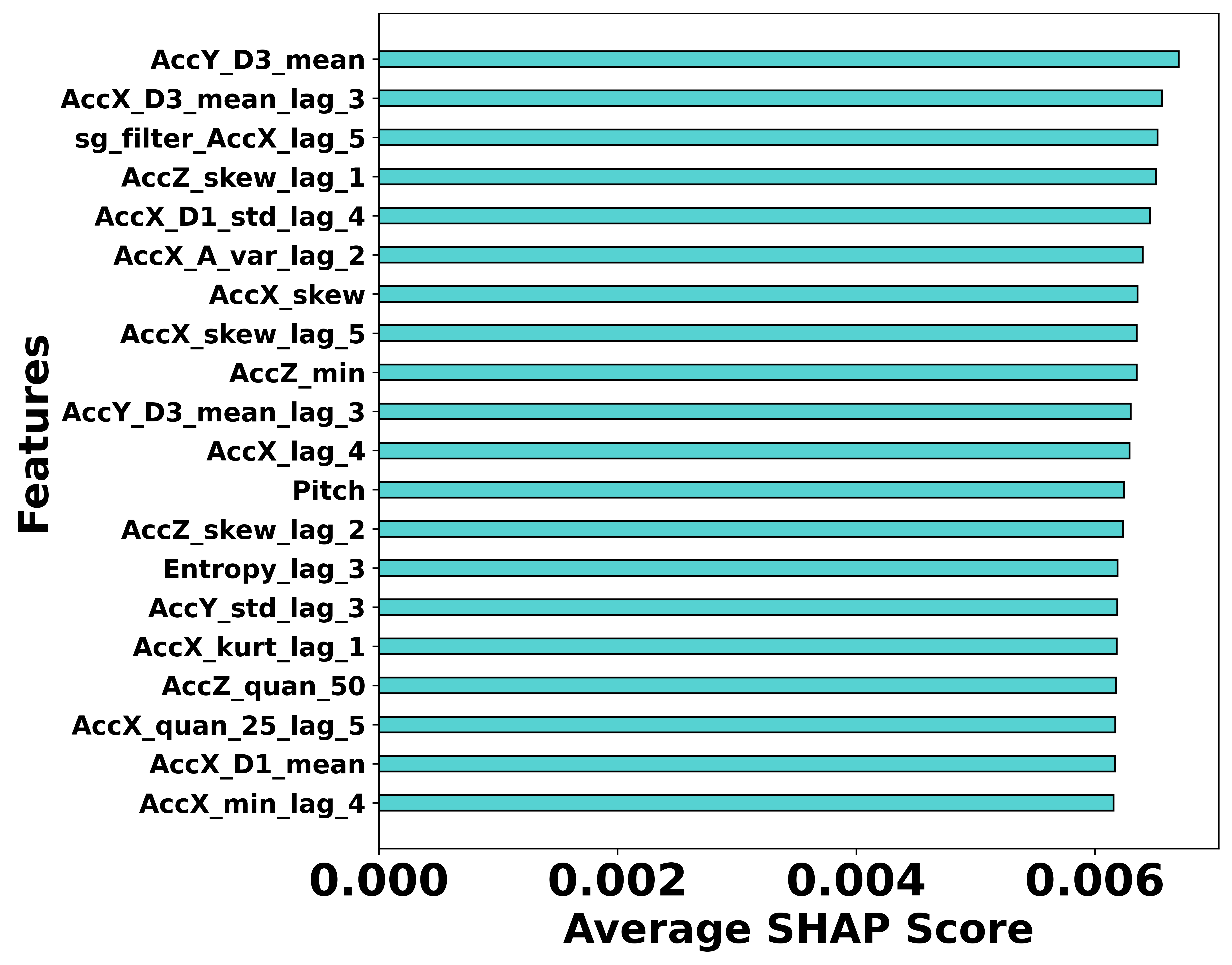}}
        \caption{} \label{fig:top_shap_values}
    \end{subfigure}
    \hfill
    \begin{subfigure}[t]{0.48\textwidth}
        \centering
        \fbox{\includegraphics[width=0.95\linewidth, height=5cm]{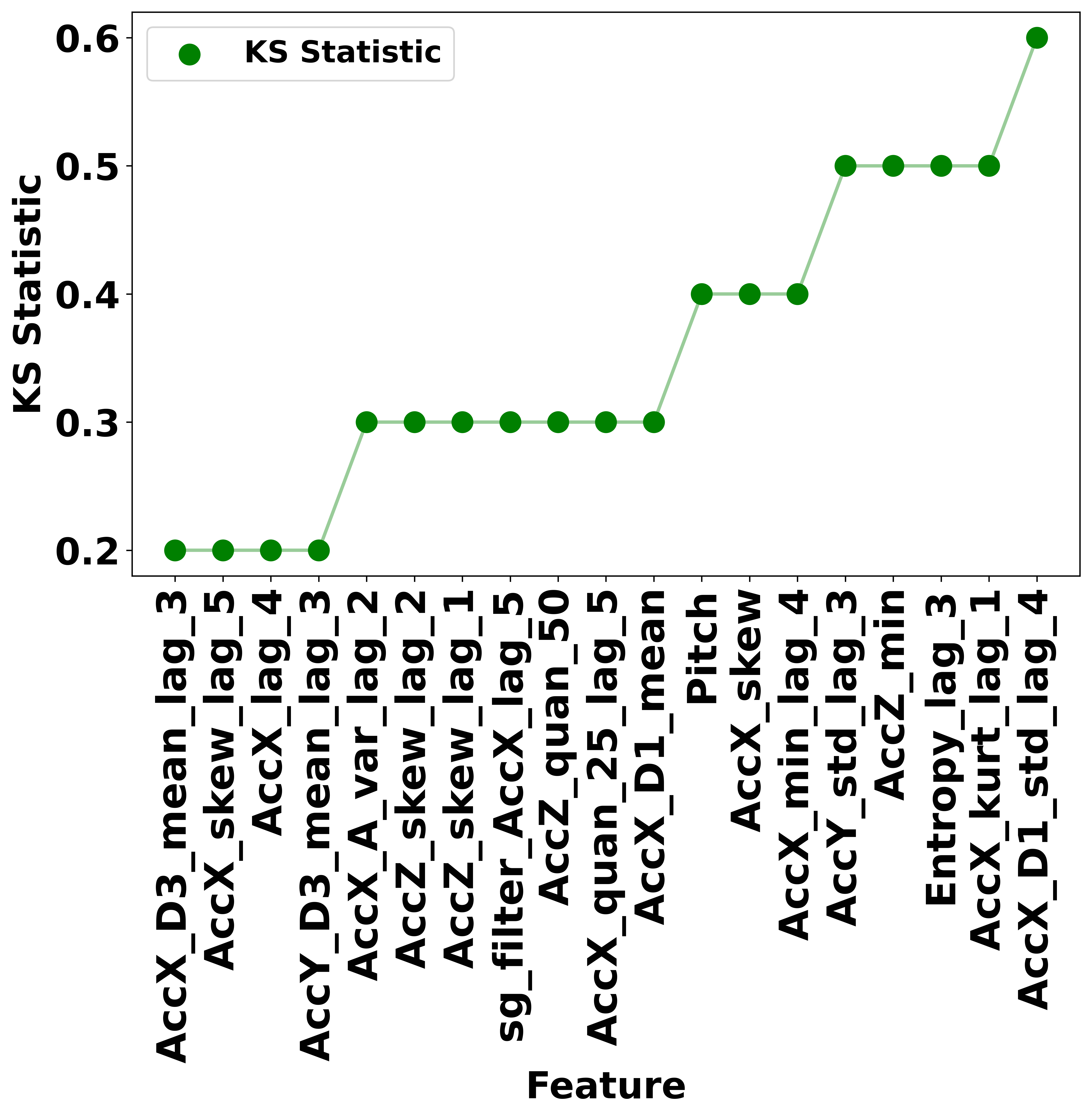}}
        \caption{} \label{fig:KS_Statistics}
    \end{subfigure}

    \caption{
    (\subref{fig:shap_bar_STN}) SHAP Analysis on Standing class
    (\subref{fig:cfm}) Confusion Matrix
    (\subref{fig:top_shap_values}) Top SHAP Features and Average SHAP Values
    (\subref{fig:KS_Statistics}) KS Statistics Results
}
    \label{fig:shap_and_cfm}
\end{figure}

\subsection{Explainable AI Implementation}

\subsubsection{SHAP Implementation}

Shapley regression values\citep{lundberg2017unified} are important features of linear models in the presence of multicollinearity. This method requires retraining the model on all feature subsets $S \subseteq F$, where $F$ is the set of all features. An importance score is assigned to each feature, reflecting its influence on the model's prediction when included. To compute this effect, a model $f_{S \cup \{i\}}$ is trained with that feature present, and another model $f_S$ is trained without it. Predictions are then made from the two models and are compared on the current input $f_{S \cup \{i\}}(x_{S \cup \{i\}}) - f_S(x_S)$, where $x_S$ represents the values of the input features in the set $S$.
Since the effect of withholding a feature depends on other features in the model, the preceding differences are computed for all possible subsets $S \subseteq F \setminus \{i\}$. The Shapley values are then computed and used as feature attributions. They are a weighted average of all possible differences:

\begin{equation}
\label{eq:shapley1}
\phi_i = \sum_{S \subseteq F \setminus \{i\}} \frac{|S|!(|F| - |S| -1)!}{|F|!} \left [ f_{S \cup \{i\}}(x_{S \cup \{i\}}) - f_S(x_S) \right ].
\end{equation}

For performing SHAP analysis, 100 values are taken randomly, 25 values from each class and then the SHAP analysis is performed.

\textbf{Standing Class:} Values of simple statistical features such as the accelerometer's X-axis's \(AccY\_D3\_mean\), \\ \(AccY\_D3\_mean\_lag\_3\), \(AccX\_D1\_std\_lag\_4\), \(AccX\_lag\_4\), \(AccX\_D3\_mean\_lag\_3\), \(AccX\_A\_std\_lag\_3\), \(AccZ\_skew\_lag\_1\), \(AccZ\_D1\_mean\_lag\_3\), \(AccX\_quan\_25\_lag\_3\), \(AccX\_A\_var\_lag\_2\) as shown in Figure \ref{fig:shap_bar_STN} contribute the most. We observe that X-axis values significantly influence the classification of the 'Standing' activity. As shown in the figure \ref{fig:cow_with_node} X-axis captures the forward and backward movement of the cows, and the Y-axis tracks the sideways movement of the head. When a cow is standing, it is moving back and forth with a little movement in neck region. We are observing that features related to those specific axes play a crucial role in detecting the cow's overall activity.

\textbf{Resting in Lying Class:} \(sg\_filter\_AccX\_lag\_5\), \(AccX\_A\_var\_lag\_2\), \(AccX\_D3\_mean\_lag\_3\), \(AccZ\_skew\_lag\_1\), \(AccY\_mean\_lag\_5\), \(AccY\_D1\_energy\), \(AccY\_sum\_lag\_2\), \(AccY\_lag\_1\), \(AccX\_A\_std\_lag\_5\), \(AccX\_skew\_lag\_5\) are the top contributor; most features from the Y-axis. For resting in lying class Y-axis is playing a major role(Supplementary Figure S1). Similarly, we have observed that the Z and X axes are also contributing, as in resting, there is a change in the Z-axis's values as it is working against gravity.

\textbf{Rumination:} \(AccY\_D3\_mean\), \(Pitch\_lag\_2\), \(AccX\_min\_lag\_4\), \(AccX\_A\_var\_lag\_2\), \(AccX\_skew\_lag\_1\), \(AccX\_mean\), \(AccX\_D3\_mean\_lag\_3\), \(sg\_filter\_AccX\_lag\_1\), \(AccX\_skew\), \(AccY\_std\_lag\_3\) are top contributors (Supplementary Figure S2). X and Y-axis values are the top important features. As for rumination activity, it mostly involves the movement of the neck. Those types of movements are best captured via the X and Y axes.

\textbf{Miscellaneous Activities:} \(AccZ\_lag\_2\), \(Pitch\), \(AccY\_A\_var\), \(AccY\_D3\_mean\), \(AccX\_skew\_lag\_4\), \(AccX\_skew\_lag\_5\), \(AccZ\_D2\_mean\), \(AccZ\_D2\_mean\_lag\_2\), \(AccX\_D2\_mean\_lag\_5\), \(AccX\_D1\_mean\) are top 10 contributors (Supplementary Figure S3). Since miscellaneous activities contain a large collection of activities so there is a presence of features from all the axes.

\begin{table}[ht]
    \centering
    \caption{Feature Scores and Stability Categories}
    \label{tab:combined_feature_table}
    \begin{tabular}{l c c l}
        \hline
        \textbf{Feature} & \textbf{Average SHAP Score} & \textbf{KS Statistic} \\
        \hline
        AccY\_D3\_mean & 0.006702 & 0.2(Moderate Stability) \\
        AccX\_D3\_mean\_lag\_3 & 0.006562 & 0.2(Moderate Stability) \\
        sg\_filter\_AccX\_lag\_5 & 0.006524 & 0.3(Moderate Stability) \\
        AccZ\_skew\_lag\_1 & 0.006510 & 0.3(Moderate Stability) \\
        AccX\_D1\_std\_lag\_4 & 0.006461 & 0.6(Instability)  \\
        AccX\_A\_var\_lag\_2 & 0.006400 & 0.3(Moderate Stability) \\
        AccX\_skew & 0.006358 & 0.4(Moderate Stability) \\
        AccX\_skew\_lag\_5 & 0.006350 & 0.2(Moderate Stability) \\
        AccZ\_min & 0.006350 & 0.5(Instability)  \\
        AccY\_D3\_mean\_lag\_3 & 0.006300 & 0.2(Moderate Stability) \\
        AccX\_lag\_4 & 0.006290 & 0.2(Moderate Stability) \\
        Pitch & 0.006246 & 0.4(Moderate Stability) \\
        AccZ\_skew\_lag\_2 & 0.006235 & 0.3(Moderate Stability) \\
        Entropy\_lag\_3 & 0.006189 & 0.5(Instability)  \\
        AccY\_std\_lag\_3 & 0.006187 & 0.5(Instability)  \\
        AccX\_kurt\_lag\_1 & 0.006182 & 0.5(Instability)  \\
        AccZ\_quan\_50 & 0.006177 & 0.3(Moderate Stability) \\
        AccX\_quan\_25\_lag\_5 & 0.006171 & 0.3(Moderate Stability) \\
        AccX\_D1\_mean & 0.006170 & 0.3(Moderate Stability) \\
        AccX\_min\_lag\_4 & 0.006156 & 0.4(Moderate Stability) \\
        \hline
    \end{tabular}
\end{table}

\subsection{Feature Stability:}

For feature stability analysis we have a subset of features which are ranked higher in SHAP  analysis Table \ref{tab:combined_feature_table} \\ \(AccX\_D1\_std\_lag\_4\), \(AccZ\_min\), \(Entropy\_lag\_3\), \(AccY\_std\_lag\_3\), \(AccX\_kurt\_lag\_1\), \(AccX\_skew\), \(Pitch\), \(AccX\_min\_lag\_4\), \(sg\_filter\_AccX\_lag\_5\), \(AccZ\_skew\_lag\_1\), \(AccX\_A\_var\_lag\_2\), \(AccZ\_skew\_lag\_2\), \(AccZ\_quan\_50\), \(AccX\_quan\_25\_lag\_5\), \(AccX\_D1\_mean\), \(AccX\_D3\_mean\_lag\_3\), \(AccX\_skew\_lag\_5\), \(AccY\_D3\_mean\_lag\_3\), \(AccX\_lag\_4\). Then we have performed KS statistics to check the stability of each feature as shown in the figure \ref{fig:top_shap_values}.

\section{Discussion}
Our solution helps in the adoption of smart agricultural solutions and sustainable practices. Cows' and overall milk production drop significantly when their health is not in good condition. Using the accelerometer's data, we can distinguish 4 classes: Standing(When a cow is standing still or moving around), REL(Resting in lying), RUS(Ruminating), and ETC(All the other miscellaneous activities). Each day, a cow does an activity for a certain amount of time, and if there is an anomaly in the ratio, we can say that the cow is having some issue with its health or surroundings. Overall, data collection and annotation were the most bandwidth-intensive tasks. We have collected and processed the data while prioritizing the consistency of the data. 3 levels of checks are implemented during the creation of the data. In Step 1, we collected and annotated the data based on video footage; in Step 2, we collected the data(ground truth) from the person present in the farm; and in the final step, we did a peer review of the annotated data so there is little to no change in data quality degradation. KNN trained with 156 window length and 39 step length is a top performer with AUROC of 0.98 $\pm$ 0.0026 (figure \ref{fig:AUC_train}), accuracy of 0.926 $\pm$ 0.0053, precision of 0.926 $\pm$ 0.0075, recall of 0.921 $\pm$ 0.0052, f1 Score 0.923 $\pm$ 0.0063 in the training set(table \ref{table:model_performance_train}) and AUROC of 0.99 (figure \ref{fig:AUC_test}), accuracy of 0.938, precision of 0.94, recall of 0.9322, f1 Score 0.936 in the testing set(table \ref{tab:model_comparison_test}). While training and testing this algorithm, we have maintained all the best practices and found a set of features. We have used SHAP-based feature selection and performed feature stability. We have observed that a set of features are coming differently, but a specific axis or set of axes dominates each activity class.

The selected features have played a key role in understanding the overall properties of an accelerometer with respect to a cow's activity. By checking the orientation of the accelerometer and the placement of the X-axis control, we can predict the forward and backward movement of the cow, which is what we are trying to predict in our solution. Using this framework, users/researchers/developers can also model and predict other activities using accelerometer data. 

\section{Conclusion}
In this study, we understand the different animal activities and the contribution of different axes to those specific activities with the help of machine learning. Both the solution and the method can be used for different use cases where the accelerometer is used as a data acquisition device. Use of statistical techniques, machine learning modeling, and interoperability frameworks has provided much-needed robustness to the solution. In the past, certain animal activities were associated with certain accelerometer axes, and it was up to us to theorize which axes contributed to each activity. With the help of this system and findings, we can have a better in-depth look at the working of the accelerometer. This study can be extended to other use cases where accelerometers are used for different readings.

\section{CRediT authorship contribution statement}
Rahul Jana: Conceptualization, Methodology, Software, Validation, Formal analysis, Investigation, Resources, Data Curation, Writing - Original Draft, Writing - review and editing, Visualisation.
Shubham Dixit: Methodology, Software, Formal analysis, Data Curation. Mrityunjay Sharma: Writing- review and editing. Ritesh Kumar: Conceptualization, Methodology, Formal analysis, Writing - review and editing, Supervision, Project administration.

\section{Declaration of competing interest}
The authors declare that they have no known competing financial interests or personal relationships that could have appeared to influence the work reported in this paper.

\section{Acknowledgement}
The author(s) acknowledge and are thankful to Council of Scientific and Industrial Research–Central Scientific Instruments Organisation (CSIR–CSIO), Sector 30, Chandigarh, India, and TIF - Agriculture and Water Technology Development Hub(AWaDH), Indian Institute of Technology, Bara Phool, Punjab - 140001, India for providing accelerometer hardware and technical support for conducting the study under the project GAP00469.

\section{Code Availability}
The complete source code used in this study is available in GitHub repository 
\href{https://github.com/CSIO-FPIL/Animal-Health-Paper}{https://github.com/CSIO-FPIL/Animal-Health-Paper}.

\section{Data Availability}
Raw and processed datasets will be made available on request.


\bibliography{references}
\newpage
\begin{center}
    \LARGE \textbf{Supplementary Information: An Explainable AI based approach for Monitoring Animal Health}
\end{center}
\renewcommand{\thefigure}{S\arabic{figure}}
\setcounter{figure}{0}  
\begin{figure}[htbp]
\renewcommand{\thefigure}{S\arabic{figure}}
\centering
\fbox{\includegraphics[width=0.7\textwidth]{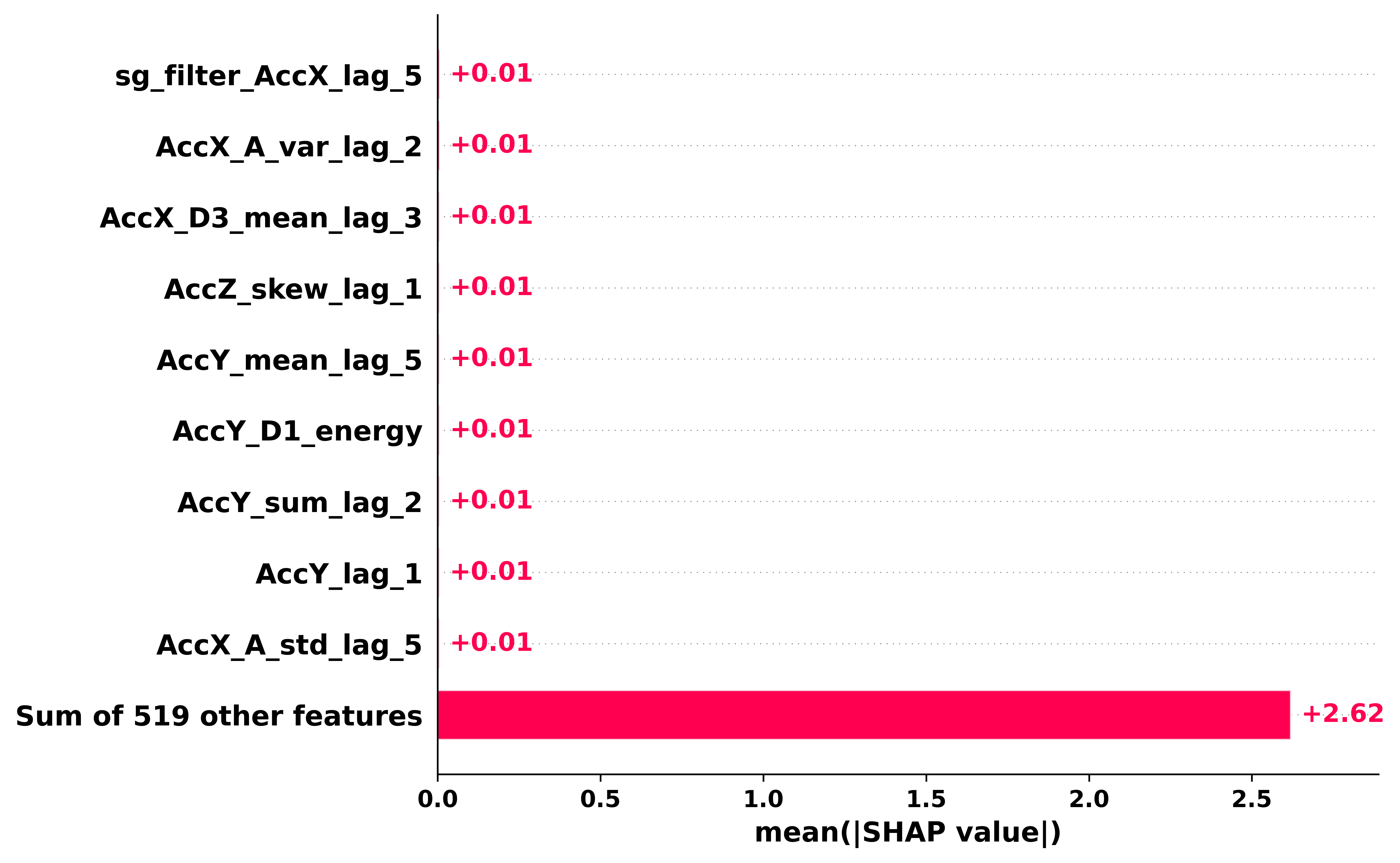}}
\caption{Average SHAP values for Resting in Lying.}
\label{fig:SHAPrel}
\end{figure}

\begin{figure}[htbp]
\renewcommand{\thefigure}{S\arabic{figure}}
\centering
\fbox{\includegraphics[width=0.7\textwidth]{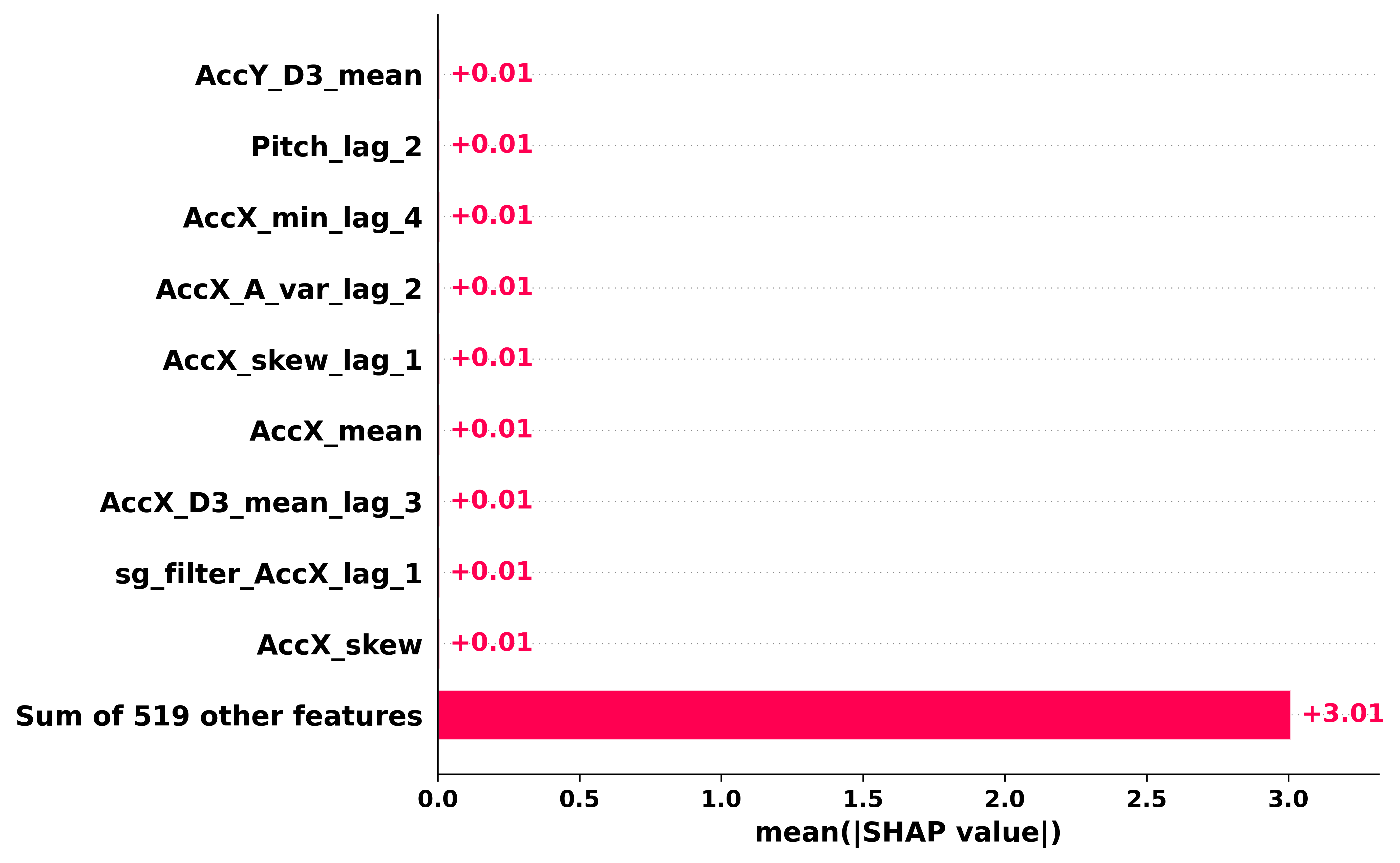}}
\caption{Average SHAP values for Rumination.}
\label{fig:SHAPrus}
\end{figure}

\begin{figure}[htbp]
\renewcommand{\thefigure}{S\arabic{figure}}
\centering
\fbox{\includegraphics[width=0.7\textwidth]{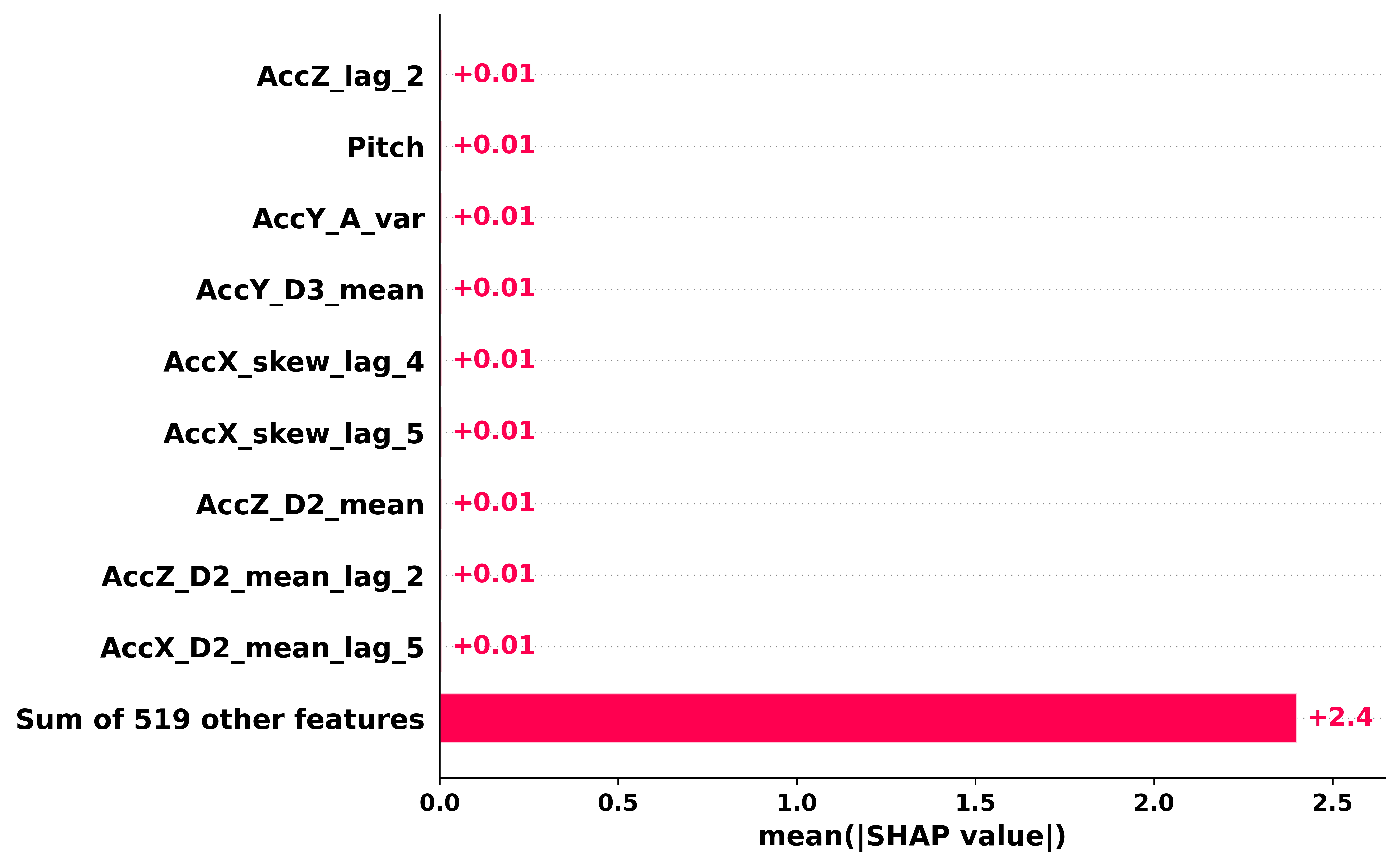}}
\caption{Average SHAP values for Miscellaneous Activities.}
\label{fig:SHAPmics}
\end{figure}

\end{document}